\documentclass{article}

\usepackage[main, final, nonatbib]{neurips_2026}

\usepackage[utf8]{inputenc} 
\usepackage[T1]{fontenc}    
\usepackage{hyperref}       
\usepackage{url}            
\usepackage{booktabs}       
\usepackage{amsfonts}       
\usepackage{nicefrac}      
\usepackage{microtype}

\usepackage[square,numbers, sort]{natbib}
\usepackage{graphicx}
\usepackage{amsmath}
\usepackage{amssymb}
\usepackage{mathtools}
\usepackage{amsthm}
\usepackage{makecell}
\usepackage[dvipsnames,svgnames,x11names,table]{xcolor}         
\usepackage[capitalize,noabbrev,nameinlink]{cleveref}

\usepackage{siunitx}
\usepackage{multirow}
\usepackage{subcaption}
\usepackage{bm}
\usepackage{wrapfig}
\usepackage{float}
\usepackage{dblfloatfix}
\usepackage{pifont}
\usepackage{enumitem}
\usepackage[most]{tcolorbox}
\definecolor{myblue}{RGB}{25,61,71}
\definecolor{myred}{HTML}{BC4B51}
\definecolor{mygreen}{HTML}{386C0B}

\tcbset{
  mycell/.style={
    colback=white,
    colframe=myblue,
    coltitle=white,
    fonttitle=\bfseries\footnotesize ,
    fontupper=\footnotesize , 
    lefttitle=0.5mm,
    leftupper=-8.75mm,
    rounded corners,
  }
}

\usepackage{caption}
\crefname{section}{section}{sections}
\crefname{figure}{figure}{figures}
\crefname{table}{table}{tables}
\crefname{appendix}{appendix}{appendices}
\crefname{equation}{equation}{equations}

\title{A Scalable Multi-Task Model for Virtual Sensors}

\author{ Leon~G\"otz$^{1\,2}$\hspace{1.8mm} Lars~Frederik~Peiss$^{1\dag}$\hspace{1.8mm} Erik~Sauer$^{1\dag}$\hspace{1.8mm} Andreas~Udo~Sass$^1$\\[0.3ex] {\bfseries Thorsten~Bagdonat$^1$\hspace{1.8mm} Stephan~G\"unnemann$^2$\hspace{1.8mm} Leo~Schwinn$^{2\,3}$} \vspace{2.0mm}\\ $^1$Volkswagen AG\hspace{2.5mm} $^2$Technical University of Munich\hspace{2.5mm} $^3$Helmholtz AI\\ $^\dag$Work done during an internship at Volkswagen AG\vspace{1.8mm}\\ Correspondence to: \texttt{leon.goetz@volkswagen.de} }

\begin{document}

\maketitle

\begin{abstract}
Virtual sensors replace expensive physical sensors in critical applications through machine learning by predicting target signals from available measurements. Existing virtual sensor approaches require application-specific models with hand-selected inputs for each sensor, cannot leverage task synergies, and lack consistent benchmarks. While emerging time series foundation models offer general-purpose, pretrained solutions in other domains, they are computationally expensive and limited to predicting their input signals, making them incompatible with virtual sensors. We introduce the first multi-task model for virtual sensors addressing both limitations. Our unified model can simultaneously predict diverse virtual sensors exploiting synergies while maintaining computational efficiency. It learns relevant input signals for each virtual sensor, eliminating expert knowledge requirements while adding explainability. In our large-scale evaluation on three standard benchmarks and an application-specific dataset with over 18 billion samples, our architecture reduces computation time by up to \SI{415}{\times} and memory requirements by \SI{951}{\times}, while maintaining or even improving predictive quality compared to unified baselines. Compared to existing isolated models for a single virtual sensor, our unified approach generates superior predictions at similar inference speed while scaling gracefully to hundreds of virtual sensors with nearly constant parameter count, enabling practical deployment in large-scale sensor networks.
\end{abstract}

\section{Introduction}
\vspace{-0.5\baselineskip}
\begin{wrapfigure}{r}{0.45\textwidth}
\vspace{-1.3\baselineskip}
    \centering
    \includegraphics[width=\linewidth,trim={0.3in 6.27in 2.46in 0.15in},clip]{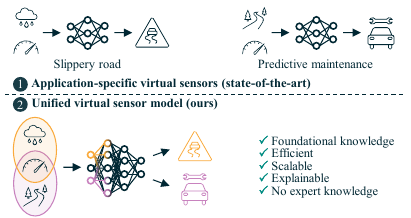}
    \vspace{-1.2\baselineskip}
    \caption{\ding{192} Recent works train, maintain, and deploy isolated application-specific models for every virtual sensor. \ding{193} We unify multiple virtual sensors into a single model, exploiting synergies and achieving superior scalability. It learns relevant input signals and selectively predicts virtual sensors, increasing efficiency and explainability without expert knowledge.}
    \label{fig:introduction}
    \vspace{-1\baselineskip}
\end{wrapfigure}
Many important applications in areas such as autonomous driving, environmental monitoring, and healthcare rely on high-quality sensors. Unfortunately, these sensors can be expensive, difficult to maintain, or unsuitable for large-scale deployment.
Virtual sensors address this limitation by computing target signals from available other measurements through statistical or learning-based methods \citep{Albertos2002virtualsensors}. They can replace physical sensors, saving costs, act as redundancy in safety-critical applications, allow for non-invasive measurements, or compute quantities that are not measurable with physical sensors, such as a battery's state of health \citep{vonbuelow2021batterysoh}. \\
\hspace{-1.3mm}While some deep learning-based approaches exist for virtual sensor simulation, they exhibit four key limitations. First, they rely on highly application-specific models that are trained, deployed, and maintained separately for each virtual sensor, with input signals typically being hand-selected by domain experts \citep{ziaukas2021sideslip}. As a result, current approaches do not scale to large sensor networks. Second, these isolated models cannot leverage synergies among different virtual sensor tasks, requiring one individual model for every simulated sensor. Third, there is little consistency in how virtual sensors are evaluated, with few works using standard benchmark datasets \citep{hu2024heartdisease}. Lastly, existing methods tend to rely on relatively simple deep learning models \citep{diniz2025oremassflow}, potentially limiting their performance.\\
Recently, foundation models have shown impressive performance in time series forecasting, outperforming specialized approaches \citep{chronos}. These unified models exploit synergies among different tasks and datasets. However, large foundation models impose considerable computation requirements \citep{goetz2025localmerging}, limiting their application in sensor networks with thousands of signals \citep{godahewa2021monash}. Additionally, current time series foundation models only forecast the input signals and cannot predict new signals, making them unsuitable for virtual sensor tasks. \\
We address these complementary gaps through a unified model for virtual sensors. By training a single model across multiple virtual sensor tasks, we overcome the scalability and synergy limitations of existing virtual sensor approaches. Simultaneously, we design our architecture to be computationally efficient, addressing the deployment constraints that prevent current foundation models from being applied to large-scale sensor networks with thousands of signals. Our key contributions are:

\textbf{Large-scale benchmark}\hspace{1.8mm}  We provide the first large-scale analysis of unified virtual sensor models on three common time series forecasting benchmarks and an application-specific automotive dataset with over \SI{17500}{\kilo\meter} of driving and \num{18}\,B time series samples using over \num{48500} H100 GPU hours.

\textbf{Unified model for virtual sensors}\hspace{1.8mm} Our unified model can predict multiple virtual sensors. By exploiting synergies among them and leveraging foundational time series knowledge, it scales gracefully to large sensor networks while even improving predictive quality in some cases. Our approach learns the set of relevant input signals for each virtual sensor, thereby adding explainability and eliminating the need for expert knowledge in virtual sensor design. Due to its flexible sensor selection mechanism, it achieves high efficiency, making it suitable for on-device computing.

\textbf{Results}\hspace{1.8mm} Our evaluation on three standard benchmarks and an application-specific automotive dataset reveals substantial computational savings while maintaining or improving performance. Our model reduces computation time up to \SI{415}{\times} and minimizes memory requirements by \SI{951}{\times} compared to standard multi-task approaches. Compared to isolated models, it generates superior predictions and scales to hundreds of virtual sensors with an almost constant number of trainable parameters.

\section{Related work}
\vspace{-0.5\baselineskip}
In this work, we design a scalable multi-task model for virtual sensor applications, aligning two previously disjoint literature branches for application-specific virtual sensors and foundation models for time series forecasting.

\textbf{Virtual sensors}\hspace{1.8mm} Virtual sensors reduce costs, improve reliability, and estimate quantities that are difficult to measure directly. In automotive applications, they estimate dynamic parameters such as the side slip angle improving safety \citep{ziaukas2021sideslip,kalyanasundaram2025sideslip}, environmentally relevant nitrogen oxide emissions \citep{arsie2017nox}, battery temperatures \citep{bamati2024batterytemperature}, and battery state of health \citep{vonbuelow2021batterysoh}. Further applications include determining ore mass flow from conveyor belt currents \citep{diniz2025oremassflow} in the mining domain, estimating moisture in food production \citep{wang2001foodmoisture}, discriminating gases in chemical processes \citep{ankara2004gasdetection}, and non-invasive measurements in healthcare \citep{filippo2023bloodpressure} to detect disease from a patient's breath \citep{hu2024heartdisease}.
While early virtual sensors used hand-crafted models \citep{Albertos2002virtualsensors}, current work predominantly learns simple models from data \citep{diniz2025oremassflow}. Nevertheless, inputs are generally selected by experts and an application-specific model is trained, deployed, and maintained for each task. These isolated models limit scalability and cannot exploit synergies among virtual sensors. Moreover, application-driven evaluations lack consistent benchmarks.

\textbf{Time series foundation models}\hspace{1.8mm} Foundation models forecast univariate \citep{das2024timesfm,chronos,rasul2024lagllama,goswami2024moment,liu2024timer,liu2025sundial,auer2025tirex} or multivariate time series \citep{woo2024unifiedtraining,gao2024units,cohen2024toto,ekambaram2024tinytimemixers,khwaja2026toto2}, optionally using covariates \citep{das2024tide,wang2024timexer,auer2025cosmic,patil2025morpheus,ansari2025chronos2}, and can outperform specialized forecasters \citep{chronos}. However, their computational cost limits deployment in sensor networks and on edge devices \citep{goetz2025localmerging}. Moreover, they require a target signal's past context and therefore cannot infer new signals solely from other measurements, as virtual sensors require. Building on the assumption of self-contained time series \citep{chen2025selfcontainedtimeseries}, some methods restrict cross-signal interactions to reduce model complexity, often estimating importance through proxies such as correlation \citep{lee2024partialcd,chen2024channelclustering,liu2024dgcformer,qiu2025duet,hu2025timefilter}, but their unstructured sparsity provides \mbox{no speedup on standard hardware.}

We propose a unified and scalable model for multiple virtual sensors within a sensor network that leverages cross-task synergies while maintaining computational efficiency. It learns the set of relevant input signals, minimizing expert knowledge requirements and enhancing explainability.  We provide a more detailed positioning of our work in \cref{appendix:positioning}.

\vspace{0.2\baselineskip}
\section{A scalable multi-task model for virtual sensors}
\label{sec:method_virtual_sensor}
\vspace{-0.5\baselineskip}
We design a unified model exploiting synergies among multiple virtual sensors. After introducing our basic model architecture, we propose a novel mechanism to selectively predict new signals that are not in the input set. We then design a mechanism to learn the set of relevant input signals per virtual sensor, minimizing expert knowledge, enhancing explainability, and improving efficiency. Finally, we introduce efficient training methods.

Let $\mathcal{Z} = \{z_i\}_{i=1}^M$ be a family of univariate real-valued time series (input signals) indexed by \mbox{$i\!=\!1, \dots, M$} and \mbox{$\mathcal{Z^\prime} = \{z^\prime_j\}_{j=1}^N$} be another family of univariate real-valued time series (virtual sensors) indexed by $j\!=\!1, \dots, N$, where $\mathcal{Z}$ and $\mathcal{Z^\prime}$ are typically disjoint but can also be overlapping, or identical. Let a neural network \mbox{$\mathbf{f}_{\boldsymbol{\theta}}(\mathbf{x}) = \mathbf{\Phi}_L \circ \mathbf{\Phi}_{L-1} \circ \cdots \circ \mathbf{\Phi}_{1}(\mathbf{x})$} with parameters $\boldsymbol{\theta}$ consist of $L$ layers denoted as $\mathbf{\Phi}_l$, where each layer takes the output of the previous layer as input. Based on a model $\mathbf{f}_{\boldsymbol{\theta}}$, we predict $N$ virtual sensors $\mathcal{Z^\prime}$ from $M$ input time series~$\mathcal{Z}$. We assume that the neural network's input $\mathbf{x}\in\mathbb{R}^{s \times d}$ consists of $s$ tokens with dimension $d$. Here, the input tokens are generated by a tokenizer $\mathbf{g}$ out of input data $\mathcal{Z}$.

\vspace{0.2\baselineskip}
\subsection{Base architecture}
\vspace{-0.3\baselineskip}
Based on the competitive results in \citet{das2024timesfm}, we use a causal decoder-only transformer as the base architecture for our model $\mathbf{f}_{\boldsymbol{\theta}}$. We tokenize our input time series $z_i \in \mathcal{Z}$ individually. To this end, we first normalize each series to have zero mean and unit standard deviation and extract non-overlapping patches of length $p$ in a second step \citep{nie2023patchtst}. Subsequently, we embed each patch into the token dimension $d$ using a multi-layer perceptron $\mathbb{R}^{p}\rightarrow \mathbb{R}^{d}$ \citep{das2024timesfm}. Finally, our tokenizer $\mathbf{g}$ concatenates all tokens into a single token sequence $\mathbf{x}\in\mathbb{R}^{s \times d}$ with time-based ordering as in \cref{fig:architecture}. 
We adopt a similar approach for de-embedding predicted tokens into time‑series representations. In the transformer, we employ a block-diagonal causal mask to impose temporal ordering without constraining interactions among time series. 

\vspace{0.2\baselineskip}
\subsection{Selective prediction of virtual sensors}
\label{sec:method_output_selection}
\vspace{-0.3\baselineskip}
We embed temporal and variate information into our tokens using sin-cos \citep{transformer} and learned position embedding, respectively. While current time series literature that forecasts only input signals omits variate information \citep{woo2024unifiedtraining,ansari2025chronos2}, embedding it explicitly is crucial for our proposed mechanism to predict virtual sensors. The temporal position embedding is shared among variates, while the variate embedding is shared orthogonally across time. We learn additive variate embedding vectors \mbox{$\mathcal{V} = \{v_i\}_{i=1}^M$} \mbox{for input signals $z_i \in \mathcal{Z}$ and $\mathcal{V^\prime} = \{v^\prime_j\}_{j=1}^N$ for virtual sensors $z^\prime_j \in \mathcal{Z^\prime}$, with~$v_i,v^\prime_j \in R^d$.}

\begin{figure*}[t]
    \centering
    \vspace{0.5\baselineskip}
    \includegraphics[width=1\textwidth,trim={0.04in 5.16in 0.44in 0.66in},clip]{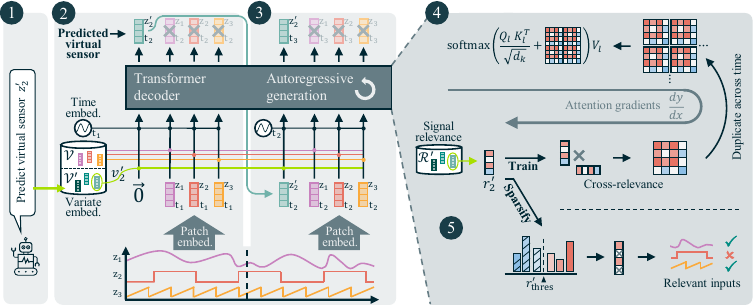}
    \vspace{-1.2\baselineskip}
    \caption{We present a unified model that can predict a user-selected target virtual sensor ($z^\prime_2$, green) out of multiple possible virtual sensors. \ding{192} We first define which virtual sensor to forecast (here sensor $z^\prime_2$). \ding{193} Next, we introduce an initially empty prototype token $\smash[t]{\vec{0}}$ and add the corresponding variate embedding $v^\prime_2 \in \mathcal{V^\prime}$ to guide the neural network to predict the specified virtual sensor, based on available sensor signals ($z_1$, pink; $z_2$, red; $z_3$, orange). We divide input signals $z_1,z_2,z_3$ into patches, embed them into tokens, and add time and variate embedding $\mathcal{V}$ to them. Starting from empty prototypes in cycle $t_1$, \ding{194} we forecast virtual sensors autoregressively using the model's previous prediction $z^\prime_2$ as current input in step $t_2$. \ding{195} Within the transformer, our architecture learns relevant input signals for each virtual sensor from attention gradients through trainable signal relevance vectors $\mathcal{R^\prime}$, adding explainability. After selecting the corresponding signal relevance vector $r^\prime_2$, we compute the cross-relevance for signal-to-signal communication, duplicate our mask across time to learn time-independent signal importance, and apply it as attention bias. \ding{196} By later sparsifying signal relevance vectors $\mathcal{R^\prime}$ based on a threshold $r^\prime_{\mathrm{thres}}$, we structurally prune irrelevant input signals to achieve considerable efficiency gains.}
    \vspace{-0.7\baselineskip}
    \label{fig:architecture}
\end{figure*}

We design a novel mechanism to forecast new signals (virtual sensors) autoregressively in a selective way.
To predict the $j$-th virtual sensor from the input signals $\mathcal{Z}$, we insert a zero-vector $\vec{0}$ as an initially empty prototype token into our token sequence $\mathbf{x}$ (see \cref{fig:architecture}). Subsequently, the additive time and variate embedding $v^\prime_j$ allows the neural network to learn the mapping of relevant target information into the prototype token. We train our model $\mathbf{f}_{\boldsymbol{\theta}}(\mathbf{x})$ to forecast the next $p$ time steps of our virtual sensor $j$ into the future. For the following autoregressive steps, we leverage our model's previous prediction as new input for virtual sensor signals~$z^\prime_j$. This way, our model starts autoregressive generation from an empty prototype token in the first step and only needs to predict residuals from increasing context in the following. 
For input signals $\mathcal{Z}$, we use ground truth measurements in all autoregressive steps. This setup mimics practical deployment scenarios where ground truth measurements from physical input sensors are available, but virtual sensor outputs must be predicted autoregressively. To forecast multiple virtual sensors in a single step, we simply add multiple prototype tokens with their respective position embeddings to the network's input. Our approach enables flexible selection of which virtual sensors to predict in each generation step, ranging from a single sensor to all sensors simultaneously. This capability addresses two practical requirements in sensor networks. First, different signals are often needed at different frequencies; for example, a vehicle's engine temperature requires higher update rates than ambient temperature due to faster control cycles. Second, by forecasting into the future, our model compensates for its own inference latency and enables predictive functionalities such as early fault detection \citep{Albertos2002virtualsensors}.

\vspace{0.2\baselineskip}
\subsection{Learning relevant input signals}
\label{method:input_subsets}
\vspace{-0.3\baselineskip}
In classical virtual sensor literature, the set of input signals $\mathcal{Z}$ is chosen by experts. However, in large sensor networks, this can be challenging and prone to error. To address this, we end-to-end learn the set of relevant input signals for each virtual sensor, minimizing expert knowledge and enhancing explainability and efficiency. To this end, we introduce signal relevance vectors $\mathcal{R^\prime} = \{r^\prime_j\}_{j=1}^N$ with $r^\prime_j \in \mathbb{R}^{M+N}$ for every virtual sensor $j$. Each vector $r^\prime_j$ encodes the relevance of all available sensors, both input signals and other virtual sensors, for predicting virtual sensor $j$, enabling hierarchical dependencies among virtual sensors. When predicting a virtual sensor, we select the corresponding relevance vector $r^\prime_j$ and compute the cross-relevance for the communication of signal pairs using the outer product $(r{^\prime_j} \cdot r{^\prime_j}^T) \in \mathbb{R}^{(M+N) \times (M+N)}$. We then duplicate our relevance matrix across time (see \cref{fig:architecture}), as we aim to learn global time-independent signal relevance. Finally, we apply this matrix $B$ in the attention score computation \citep{transformer} of every transformer layer $l$ as static bias in \cref{eq:attention_score}. At the beginning of training, we initialize our relevance vectors $\mathcal{R^\prime}$ with ones, as zero vectors would prevent gradient flow, utilizing the translation invariance of the softmax function.
\vspace{0.4\baselineskip}
\begin{equation}
\mathrm{Attention}(Q, K, V)_l = \mathrm{softmax}\left(\frac{Q_l K^T_l}{\sqrt{d_k}} + B\right) V_l
\label{eq:attention_score}
\vspace{0.4\baselineskip}
\end{equation}
While current work determines signal importance using proxy measures such as correlation at the time series level \citep{lee2024partialcd}, we directly learn relevance by backpropagating gradients from attention score computation to our relevance vectors $\mathcal{R^\prime}$. As we share our mask across transformer layers, signal relevance is determined based on simple features in the first layers, but can leverage more complex representations in deeper layers. Note that our relevance mechanism is not designed to learn complex inter-signal dynamics, which are modeled by the transformer, but to identify the general set of input signals relevant to a given virtual sensor.
\newpage
\textbf{Efficient processing via sparsification}\hspace{1.8mm} To increase the efficiency of our unified model and enable applications on edge devices, we sparsify our relevance vectors $r^\prime_j \in \mathcal{R^\prime}$ at late training stages and inference time. To this end we mask unimportant signals in $r{^\prime_j}$  with $-\infty$ based on a threshold\footnote{Relevance is learned as soft, differentiable scores before hard thresholding removes already-downweighted signals without destabilizing optimization.} $r^\prime_{\mathrm{thres}}$, which controls the sparsity of the attention map and which signals are used for forecasting. Our masking is equivalent to completely disregarding unimportant signals as the outer product operation and duplication across time generate fully masked rows and columns in our attention bias $B$ in a structured way. This enables us to learn individual input signal sets $\{\mathcal{Z}_j \subseteq \mathcal{Z}\}_{j=1}^N$ for every virtual sensor. Together with our virtual sensor selection mechanism in \cref{sec:method_output_selection}, we achieve maximum efficiency by querying our model only for currently requested virtual sensors and using only relevant input signals for these sensors. This reduces the number of tokens $s$ to be processed to a minimum. When requesting multiple virtual sensors simultaneously, we simply average the corresponding attention biases $B$ and union the individual input signal sets.

\subsection{Efficient training strategies}
\vspace{-0.5\baselineskip}
We optimize the parameters $\boldsymbol{\theta} \in \boldsymbol{\Theta}$ of our model $\mathbf{f}_{\boldsymbol{\theta}}$ to minimize the mean squared error only of virtual sensor forecasts. Predicted input signals are not included in loss computation (see \cref{fig:architecture}). Training all virtual sensors $z^\prime_j \in \mathcal{Z^\prime}$ simultaneously in every training step, however, is impractical as all signal relevance vectors $r{^\prime_j} \in \mathcal{R^\prime}$ would be updated with identical gradients, resulting in identical input sets $\mathcal{Z}_j \subseteq \mathcal{Z}$ for all sensors\footnote{Adding random noise mitigates identical gradients but cannot ensure the correct mapping of input signals to virtual sensors.}. Training a single virtual sensor at a time, instead, is inefficient, prolonging the training process due to less informative gradients. As a consequence, we select a random subset of $N_{\mathrm{train}} < N$ virtual sensors in every training iteration. Similar to dropout \citep{Srivastava2014dropout}, this reduces the correlation among relevance vectors. 
\vspace{1pt}

\textbf{Teacher forcing}\hspace{1.8mm} Our model generates virtual sensor predictions autoregressively as described in \cref{sec:method_output_selection}. Standard training via backpropagation through time \citep{rumelhart1985backpropagation} would incur high memory and computational costs due to long recurrent gradient paths and multiple forward passes per training step.
Thus, we employ teacher forcing \citep{williams1989teacherforcing}, removing the recurrent path in \cref{fig:architecture}, and use ground truth data for virtual sensors during training. This enables simultaneous training on multiple context lengths with a single model iteration, yielding short gradient paths and fast training cycles. However, teacher forcing limits the model’s ability to learn error propagation strategies. To mitigate this, we explore hybrid training \mbox{schedules combining teacher forcing and backpropagation through time in \cref{results:efficient_training_new}.}

\section{Experiments}
\label{sec:experiments}
\vspace{-0.5\baselineskip}
We systematically train and evaluate our unified model on \num{16} virtual sensor tasks and \num{4} datasets, demonstrating performance and efficiency gains of our scalable architecture. In \cref{appendix:experiments}, we provide further experimental details.

\textbf{Datasets}\hspace{1.8mm} We use Traffic, Electricity, and Solar as standard time series forecasting benchmarks \citep{godahewa2021monash}.
For application-specific evaluation in a realistic deployment scenario we use an automotive CAN bus dataset.
The CAN bus (controller area network bus) enables communication between electronic control units in vehicles \citep{ISO11898}. Our dataset contains \num{1713} variates including continuous sensor measurements, categorical states, and event-based signals. It features substantially more complex signal interactions compared to the standard time series datasets, e.g., the categorical driving mode \mbox{(eco, sport, comfort, ...)} influences pedal curves, stability, power delivery, efficiency, and generally driving behavior. Our CAN bus dataset is recorded with an Audi e-tron electric vehicle covering \num{287} hours of driving and over \SI{17500}{\kilo\meter} on German roads. It is synchronized to \SI{100}{\milli\second} granularity and over \SI{1000}{\times} larger than the other datasets with approximately \num{18}\,B time series samples.

\textbf{Virtual sensors}\hspace{1.8mm} 
For the Traffic, Electricity, and Solar datasets, we randomly select our set $\mathcal{Z^\prime}$ of $N=16$ virtual sensors, as all variates present the same quantity, measured at different locations. In the CAN bus dataset, every signal has a distinct purpose. Here, we hand-pick $N=16$ virtual sensors with clear intuitive interpretations, including electrical powertrain voltages, currents, and temperatures, coolant intake and outflow temperatures and flow-rates, torques and velocities.

\textbf{Models}\hspace{1.8mm} Throughout our experiments we train and evaluate models with $L=4$ transformer layers, \num{4} heads, and token dimension $d=512$. All our models and baselines have similar capacity with \num{7.4}\,M trainable parameters for Traffic, \num{7.2}\,M for Electricity, \num{7.1}\,M for Solar, and \num{7.8}\,M  for the CAN bus dataset, due to different numbers of learned variate embedding vectors $\mathcal{V}$. The signal relevance vectors $\mathcal{R^\prime}$ account for only \num{13.8}\,k, \num{5.1}\,k, \num{2.2}\,k, and \num{27.4}\,k parameters, respectively. We utilize patches with length $p=32$ to divide time series into tokens. For evaluation, we perform \num{6} autoregressive generation cycles. For LSTM-based literature models, we likewise employ \num{4} layers with a hidden dimension of \num{512}, yielding strong baselines with a larger number of parameters.

\section{Results}
\vspace{-0.5\baselineskip}
\crefname{section}{sec.}{secs.}
\crefname{appendix}{app.}{app.}
The overarching goal of our experiments is to investigate the feasibility of a unified virtual sensor model. First, we demonstrate efficiency and superior scaling properties (\cref{sec:main_experiment}). In particular, we investigate efficiency gains through focusing only on relevant input signals (\cref{sec:input_sparsity}) and through flexibly predicting a subset of virtual sensors (\cref{sec:output_selection}). We then explore the scaling of our unified model to hundreds of virtual sensors (\cref{sec:results_scaling}) and highlight explainability benefits of our architecture (\cref{sec:explainability}). Finally, we investigate \mbox{efficient training techniques (\cref{results:efficient_training_new}) including transfer learning (\cref{appendix:transfer_learning}).}
\crefname{section}{section}{sections}
\crefname{appendix}{appendix}{appendices}

\begin{table*}[b]
    \small
      \caption{Comparison of a simple \textbf{unified model} predicting all \num{16} virtual sensors, where we subsequently introduce our proposed mechanisms, with \num{16} \textbf{individual models} predicting a single virtual sensor each. We list MSE, inference time, dynamic CUDA peak memory, and the number of trainable parameters for Traffic and CAN bus datasets.}
      \vspace{-0.5\baselineskip}
      \label{tab:main_results}
  \centering
  \resizebox{\textwidth}{!}{
  \begin{tabular}{lrrrr@{\hspace{1.0cm}}rrrr}
    \toprule
    \multirow{2}{*}{Architecture} & \multicolumn{4}{@{\hspace{-0.7cm}}c}{Traffic} & \multicolumn{4}{c}{CAN bus}\\\cmidrule(r{0.95cm}){2-5} \cmidrule(l{0.0em}){6-9}
     & MSE & Time & Mem. & Param. & MSE & Time & Mem. & Param. \\
     \midrule
    \textbf{Individual models}&&&&&&&&\\
    LSTM (literature)  & \textcolor{myred}{\num{0.316}} & \textcolor{mygreen}{\SI{0.33}{\milli\second}} & \textcolor{mygreen}{\SI{0.0010}{\giga\byte}} & \textcolor{myred}{\num{145.9}\,M} & \textcolor{mygreen}{\num{0.124}} & \textcolor{mygreen}{\SI{0.35}{\milli\second}} & \textcolor{mygreen}{\SI{0.0011}{\giga\byte}} & \textcolor{myred}{\num{174.7}\,M}\\
    Transformer (ours) & \textcolor{mygreen}{\num{0.285}} & \textcolor{myred}{\SI{16.62}{\milli\second}} & \textcolor{myred}{\SI{0.90}{\giga\byte}} & \textcolor{myred}{\num{118,4}\,M} & \textcolor{myred}{\num{0.318}} & \textcolor{myred}{\SI{58.05}{\milli\second}} & \textcolor{myred}{\SI{3.42}{\giga\byte}} & \textcolor{myred}{\num{124.8}\,M}\\
    \midrule
    \textbf{Unified model} & \textcolor{myred}{\num{0.556}} & \textcolor{myred}{\SI{16.70}{\milli\second}} & \textcolor{myred}{\SI{0.90}{\giga\byte}} & \textcolor{mygreen}{\num{7.0}\,M} & \textcolor{myred}{\num{0.727}} & \textcolor{myred}{\SI{58.19}{\milli\second}} & \textcolor{myred}{\SI{3.46}{\giga\byte}} & \textcolor{mygreen}{\num{7.0}\,M}\\
    +\,Variate embedding $\mathcal{V},\mathcal{V^\prime}$ & \textcolor{mygreen}{\num{0.275}} & \textcolor{myred}{\SI{16.76}{\milli\second}} & \textcolor{myred}{\SI{0.90}{\giga\byte}} & \textcolor{mygreen}{\num{7.4}\,M} & \textcolor{mygreen}{\num{0.119}} & \textcolor{myred}{\SI{58.23}{\milli\second}} & \textcolor{myred}{\SI{3.46}{\giga\byte}} & \textcolor{mygreen}{\num{7.8}\,M}\\[0.5em]
    \makecell[l]{+\,Signal relevance $\mathcal{R^\prime}$\\+\,$N_{\mathrm{train}}=4$} & \textcolor{mygreen}{\num{0.272}} & \textcolor{myred}{\SI{16.85}{\milli\second}} & \textcolor{myred}{\SI{0.90}{\giga\byte}} & \textcolor{mygreen}{\num{7.4}\,M} & \textcolor{mygreen}{\num{0.136}} & \textcolor{myred}{\SI{58.41}{\milli\second}} & \textcolor{myred}{\SI{3.46}{\giga\byte}} & \textcolor{mygreen}{\num{7.8}\,M}\\[1em]
    \makecell[l]{+\,Sparse input sets $\mathcal{Z}_j$\\+\,Sensor selection} & \textcolor{mygreen}{\num{0.282}} & \textcolor{mygreen}{\SI{0.87}{\milli\second}} & \textcolor{mygreen}{\SI{0.0352}{\giga\byte}} & \textcolor{mygreen}{\num{7.4}\,M} & \textcolor{mygreen}{\num{0.121}} & \textcolor{mygreen}{\SI{0.14}{\milli\second}} & \textcolor{mygreen}{\SI{0.0036}{\giga\byte}} & \textcolor{mygreen}{\num{7.8}\,M}\\
    \bottomrule
  \end{tabular}
  }
  \vspace{-0.6\baselineskip}
\end{table*}

\subsection{Main experiments}
\label{sec:main_experiment}
\vspace{-0.5\baselineskip}
The application-specific nature of virtual sensor methods has resulted in a lack of consistent benchmarks. Meanwhile, time series foundation models cannot predict new, unseen signals from other measurements, limiting their applicability to virtual sensor tasks (see \cref{appendix:positioning} for a detailed comparison). We address both gaps through two contributions. First, we establish a baseline on standardized benchmarks. Here, we train an individual model for every sensor as done in prior work \citep{ziaukas2021sideslip}. Second, we introduce a unified baseline that predicts all virtual sensors from available input signals at every step and systematically introduce and ablate novel components that enable the training and deployment of a large-scale multi-task model: (i) variate embeddings, \mbox{(ii) learned relevance masking, (iii) sparsification, and (iv) sensor selection.}

\textbf{Individual models}\hspace{1.8mm} Following state-of-the-art approaches from literature, we train \num{16} isolated models to predict a single virtual sensor each. Prior works predominantly rely on standard recurrent architectures \citep{ziaukas2021sideslip,arsie2017nox,bamati2024batterytemperature} and we thus implement LSTMs \citep{hochreiter1997lstm} as a baseline. 
For an architecture-independent evaluation, we additionally train individual models with our transformer architecture. Our unified approach outperforms all literature baselines and is superior in \num{3} out of \num{4} settings in our atomic comparison in \cref{tab:main_results,tab_appendix:main_results}, demonstrating its effectiveness. Remarkably, our unified model for \num{16} virtual sensors achieves comparable average inference speed and, in some cases, even exceeds the efficiency of isolated models trained for a single virtual sensor. Moreover, training our models using backpropagation through time, analogous to LSTM-based virtual sensors, further reduces the MSE to \num{0.083} on the CAN bus dataset in \cref{tab:bptt_training}, improving considerably over the LSTM baseline. 
Relying on foundational time series understanding, our multi-task model scales superior compared to training, maintaining, and deploying \num{16} isolated models with in total \SI{16}{\times} the number of parameters, which we further investigate in \cref{sec:results_scaling}.

\textbf{Unified models}\hspace{1.8mm} We systematically introduce our novel mechanisms in a simple unified baseline. \textbf{(i)} Our results in \cref{tab:main_results,tab_appendix:main_results} demonstrate that variate embeddings are crucial for predicting virtual sensors. This is contrary to current time series forecasting literature that can only predict input signals  \citep{woo2024unifiedtraining,ansari2025chronos2}. \textbf{(ii)} Further, introducing signal relevance vectors and training only a subset of virtual sensors per iteration $N_{\mathrm{train}} / N = 4/16$ adds explainability and does not significantly affect predictive quality or efficiency. \textbf{(iii)} Additionally, we sparsify our learned signal relevance to determine individual input signal subsets $\mathcal{Z}_j \subseteq \mathcal{Z}$ for each virtual sensor. \textbf{(iv)} Together with our sensor selection mechanism, where we flexibly query our model to predict only currently required virtual sensors, we achieve high efficiency gains. Remarkably, we improve inference time by \SI{415}{\times} and reduce memory requirements by \SI{951}{\times} without affecting MSE on the CAN bus dataset. This enables our unified model to run on edge devices in large sensor networks. In \cref{appendix:main_experiments} we present results for Electricity and Solar.

\subsection{Different MSE efficiency trade-offs for variable test-time computation}
\label{sec:input_sparsity}
\vspace{-0.5\baselineskip}
\begin{wrapfigure}{r}{5.4cm}
    \centering
    \vspace{-1.0\baselineskip}
    \includegraphics[width=5.4cm,trim={0in 0.0in 0in 0in},clip]{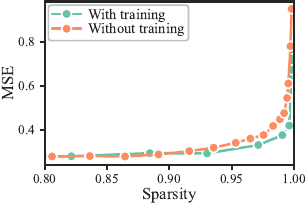}
    \vspace{-1.5\baselineskip}
    \caption{Varying our signal importance threshold $r^\prime_{\mathrm{thres}}$ during training or inference generates smooth trade-offs between input set sparsity and MSE on the Traffic dataset.}
    \label{fig:mse_efficiency_inputsets}
    \vspace{-1.0\baselineskip}
\end{wrapfigure}
To distinguish between important input signals $\mathcal{Z}_j \subseteq \mathcal{Z}$ and unimportant ones for each virtual sensor, we threshold our signal relevance vectors $r^\prime_j \in \mathcal{R^\prime}$ with $r^\prime_{\mathrm{thres}}$. Varying this threshold, we achieve different trade-offs between predictive quality and efficiency. Sparsification towards the end of training allows the model to adapt to reduced input sets, while sparsification only during inference might be more flexible. To compare both approaches, we train models with different sparsity thresholds. \\
Our results in \cref{fig:mse_efficiency_inputsets} show that we can achieve high input sparsity without affecting predictive quality as most input signals are unimportant for individual virtual sensor tasks. Towards very high sparsity, MSE increases as we now also remove informative signals from the input sets. The approach is highly flexible in practice, as sparsification can be applied solely at inference time with minimal performance degradation. On the Traffic dataset, models trained with sparse inputs achieve only marginal improvements over inference-time sparsification. Moreover, on CAN bus data, both sparsification approaches exhibit similar performance (see \cref{sec:appendix_mse_efficiency}). This allows inference-time sparsification to flexibly adapt to varying computational budgets for each prediction, without requiring the loading of different model checkpoints. Only the set of input signals is dynamically adjusted. This is particularly relevant in real-world applications where computational resources may be limited during critical situations, or where faster inference is required to ensure timely decision-making.

\subsection{Sensor selection maximizes efficiency without affecting predictive quality}
\label{sec:output_selection}
\vspace{-0.5\baselineskip}
\begin{wrapfigure}{r}{5.4cm}
    \centering
    \vspace{-1.0\baselineskip}
    \includegraphics[width=5.4cm,trim={0in 0.0in 0in 0in},clip]{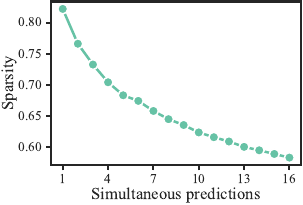}
    \vspace{-1.5\baselineskip}
    \caption{Sparsity increases as fewer virtual sensors are predicted simultaneously through our sensor selection mechanism. Results are shown for the Traffic dataset.}
    \label{fig:output_selection_traffic}
    \vspace{-1.0\baselineskip}
\end{wrapfigure}
In practical applications, virtual sensors are often required at varying frequencies. For example, engine-related measurements typically demand higher sampling rates than ambient signals in vehicles. Moreover, different virtual sensors generally rely on distinct sets of relevant input signals. To predict multiple virtual sensors simultaneously, we union their respective input signal sets, reducing sparsity. In the limiting case of equally sized, disjoint input sets, sparsity decreases linearly with the number of predicted virtual sensors. To address this, we introduce a sensor selection mechanism that enables the model to predict only the virtual sensors currently required, ensuring maximum efficiency when a specific signal is requested. In the following, we demonstrate the effectiveness of our approach by predicting an increasing number of virtual sensors simultaneously. 
By selectively predicting subsets of virtual sensors -- ranging from all sensors to individual sensors -- per autoregressive step, we increase sparsity from \SI{58.4}{\percent} to \SI{81.9}{\percent} on the Traffic dataset (see \cref{fig:output_selection_traffic}). This translates to an additional \SI{4.8}{\times} inference acceleration and \SI{4.6}{\times} memory reduction. For CAN bus data efficiency gains are even higher with \SI{7.9}{\times} and \SI{13.4}{\times}, respectively (see \cref{sec:appendix_sensor_selection}). As more virtual sensors are predicted simultaneously, sparsity decreases sub-linearly since related sensors often share common input signals. The number of simultaneously predicted signals does not affect predictive quality in our experiments.

\subsection{Scaling to hundreds of virtual sensors}
\label{sec:results_scaling}
\vspace{-0.5\baselineskip}
Classical approaches train a single model for each virtual sensor. As the number of virtual sensors grows, this causes a linear increase in training and deployment effort and model parameters, making it impractical at scale.
In contrast, our unified model maintains an almost constant number of trainable parameters when scaling to large sensor networks. However, modeling capacity might be a limitation for our fixed-size neural network. To investigate this, we train models to predict increasing numbers of virtual sensors $N_{\mathrm{Traffic}} \in \{4, 8, 16, ..., 512\}, N_{\mathrm{CAN}} \in \{4, 8, 12, 16\}$ and evaluate them on a constant set of \num{4} core sensors. \\
As the number of learned virtual sensors increases, predictive quality of our models remains stable with a standard deviation of \num{0.028} on Traffic and \num{0.008} on the CAN bus dataset. This demonstrates that our multi-task model possesses sufficient capacity to learn hundreds of virtual sensors. We argue that our architecture scales gracefully by leveraging shared knowledge across virtual sensors.

\subsection{Explainability of virtual sensors}
\label{sec:explainability}
\vspace{-0.5\baselineskip}
Beyond substantial efficiency improvements, learning the relevant input signal sets for each virtual sensor enhances explainability, which is particularly important in safety-critical applications. Furthermore, it allows for robustness analysis to identify which virtual sensors remain functional under conditions of incomplete input availability. Here, we investigate the learned signal relevance in detail.

\textbf{Qualitative assessment}\hspace{1.8mm} We utilize expert knowledge to analyze the learned input signal sets for the individual virtual sensors. For this purpose, we use the CAN bus dataset, as every virtual sensor can be associated with an interpretable physical quantity. We find that our relevance vectors learn meaningful selections. Virtual sensors for temperatures mainly rely on  other temperatures together with coolant flow rates, while virtual sensors for torques utilize engine speeds, velocities, and acceleration signals. Sensors for electrical voltages, in contrast, rely on other voltages and electrical current signals as we visualize in \cref{appendix:learning_input_sets}. This facilitates the benefit of learning individual input sets for each virtual sensor. Overall, the input sets \mbox{align well with domain expert assessments, confirming their plausibility.}

\begin{wraptable}{r}{0.45\textwidth}
      \vspace{-1.1\baselineskip}
      \caption{Comparison of different methods to determine signal relevance on the Traffic and the CAN bus dataset. \textbf{Best} in bold.}
      \label{tab:quantitative_assessment}
      \vspace{-0.5\baselineskip}
      \centering
        \resizebox{1\linewidth}{!}{
        \begin{tabular}{lrrrr}
            \toprule 
            \multirow{2}{*}{Signal relevance} & \multicolumn{2}{c}{Traffic} & \multicolumn{2}{c}{CAN bus}\\\cmidrule{2-3}\cmidrule{4-5}
             & Sparsity & MSE & Sparsity & MSE\\ 
            \midrule 
            Random & \num{0.820} & \num{0.319} & \num{0.977} & \num{0.601}\\
            Correlation & \num{0.820} & \num{0.492} & \num{0.978} & \num{0.267}\\
            SHAP & \num{0.820} & \num{0.311} & \num{0.978} & \num{0.209}\\
            Learned & \num{0.821} & \textbf{0.282} & \num{0.978} & \textbf{0.121}\\
            \bottomrule 
        \end{tabular}
        }
    \end{wraptable}
\textbf{Quantitative assessment}\hspace{1.8mm} We learn input signal relevance through attention gradients. Here, we compare our approach to choosing input signals randomly, utilizing signal correlations between inputs and virtual sensors as a proxy metric, and using SHAP \citep{lundberg2017shap} as a post-hoc relevance attribution method. For a fixed sparsity, our approach substantially outperforms random, correlation-, and SHAP-based input signal sets as our results in \cref{tab:quantitative_assessment} show. We argue that learning relevant signals directly through attention gradients exploits both semantically simple features and complex ones in deeper neural network layers. In contrast, correlation as a proxy only relies on simple time series features, which may represent a considerable limitation for complex sensor networks, such as the CAN bus. Further, end-to-end learning relevance throughout the training process improves over post-hoc attribution via SHAP.

\begin{wrapfigure}{r}{5.4cm}
    \centering
    \vspace{-1.0\baselineskip}
    \includegraphics[width=5.4cm,trim={0in 0.0in 0in 0in},clip]{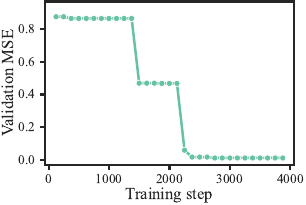}
    \vspace{-1.5\baselineskip}
    \caption{Our architecture subsequently identifies \num{2} virtual sensors among \num{1000} signals in our synthetic random dataset during training, reducing the validation loss in a stepwise manner.}
    \label{fig:toy_example}
    \vspace{-1.0\baselineskip}
\end{wrapfigure}
\textbf{Synthetic data}\hspace{1.8mm} Our model learns to identify input signals based on their variate embedding, to attribute importance scores to them through signal relevance vectors, and to compute the right virtual sensor according to our sensor selection mechanism. Here, we validate the seamless interplay of these three mechanisms, leveraging synthetic data. To this end, we generate a dataset with $M=1000$ random, uncorrelated signals. We train a model to replicate $N=2$ input signals $z_{100}, z_{200}$ as virtual sensors $z^\prime_1, z^\prime_2$, respectively. Throughout the learning process, our model subsequently identifies both input signals (see \cref{fig:toy_example}). It attributes at least \SI{4.22}{\times} higher relevance to them compared to all other \num{998} signals. Finally, it selects the correct signal as virtual sensor output. This experiment demonstrates the seamless interaction of identification, relevance, and selection mechanisms in our architecture.

\subsection{Efficient training strategies}
\label{results:efficient_training_new}
\vspace{-0.5\baselineskip}

\begin{wrapfigure}{r}{5.4cm}
    \centering
    \vspace{-1.0\baselineskip}
    \includegraphics[width=5.4cm,trim={0in 0.0in 0in 0in},clip]{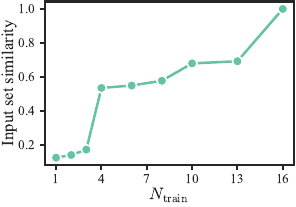}
    \vspace{-1.5\baselineskip}
    \caption{Learning different numbers of virtual sensors $N_{\mathrm{train}}$ in a single training iteration influences the similarity of their input signal sets on the Traffic dataset.}
    \label{fig:tv_per_sample_train_traffic}
    \vspace{-0.5\baselineskip}
\end{wrapfigure}
\textbf{Training virtual sensors simultaneously}\hspace{1.8mm} Learning multiple virtual sensors per training iteration improves training efficiency. However, training all virtual sensors in every cycle ultimately results in identical signal relevance vectors, as they receive identical gradients and therefore identical input signal sets for all virtual sensors. We explore this trade-off in detail by training models with $N_{\mathrm{train}} \in \{1,2,3,4,6,8,10,13,16\}$ and measure the average similarity of the sensors' input signal sets $\mathcal{Z}_j \subseteq \mathcal{Z}$. \\
Training a single virtual sensor at a time results in most specialized input sets that share \SI{12.7}{\percent} of the signals on average as in \cref{fig:tv_per_sample_train_traffic}. However, training generally requires more iterations due to less informative gradients. In contrast, learning multiple virtual sensors simultaneously results in more similar input signal sets, but makes training more efficient. In practice, we find $N_{\mathrm{train}} = 4$ to be a good tradeoff between training time and input set separation. We present CAN bus \mbox{dataset results in \cref{sec:appendix_simult_training} and additional analysis in \cref{sec_appendix:interactions}.}

\vspace{0.5\baselineskip}

\begin{wraptable}{r}{0.45\textwidth}
      \vspace{-1.1\baselineskip}
      \caption{Comparison of teacher forcing, back propagation through time (BPTT), and a combination of both for training our model on the CAN bus dataset. We list MSE, time per training iteration, and dynamic CUDA peak memory. \textbf{Best} in bold.}
      \label{tab:bptt_training}
      \vspace{-0.5\baselineskip}
      \centering
        \resizebox{1.0\linewidth}{!}{
        \begin{tabular}{lcrr}
    \toprule
    Training strategy & MSE & Time & Mem.  \\
    \midrule
            Teacher forcing & \num{0.136} & \textbf{0.19\thinspace s} & \textbf{16.6\thinspace GB}\\
            BPTT & \ding{55} & \SI{1.11}{\second} & \SI{72.0}{\giga\byte}\\
            BPTT fine tuning  & \textbf{0.083} & \SI{1.10}{\second} & \SI{72.0}{\giga\byte}\\
            \bottomrule 
        \end{tabular}
        }
    \end{wraptable}
\textbf{Teacher forcing}\hspace{1.8mm} Our model generates virtual sensor predictions autoregressively during inference. For training, however, we utilize teacher forcing as a more efficient strategy. Compared to backpropagation through time, teacher forcing requires only a single model forward pass, substantially reducing memory usage and training time. However, this approach prevents the model from learning to correct for accumulated errors during inference.
We compare both training strategies in the following.\\
Even with extensive hyperparameter tuning, training with backpropagation through time diverges. We argue that this is due to long recurrent gradient paths. Conversely, training with teacher forcing converges smoothly, while being \SI{5.8}{\times} faster and consuming \SI{4.3}{\times} less memory, as shown in \cref{tab:bptt_training}. Throughout our paper, the low inference MSE indicates that models trained with teacher forcing successfully predict initial tokens from empty prototypes, as well as subsequent tokens, and do not suffer from extensive error accumulation in autoregressive forecasting. We do not find a significant difference when comparing the predictive loss of initial tokens to that of subsequent ones. We provide further evidence for the effectiveness of teacher forcing in \cref{appendix:visualization}.
Additionally, we experiment with backpropagation through time as a fine-tuning step starting from a teacher-forcing checkpoint. This hybrid schedule allows us to maintain efficient training for most of the process while explicitly introducing error propagation toward the end of training. \Cref{tab:bptt_training} shows that this approach further improves predictive performance. 

\textbf{Transfer learning}\hspace{1.8mm} We explore transfer learning to introduce a new virtual sensor into an already trained model with limited data to explore the generalization ability of our unified approach. Our results in \cref{appendix:transfer_learning} show that transfer learning achieves the same MSE as training from scratch, while only using \SI{10}{\percent} of training data and \SI{3.6}{\percent} of computation.

\section{Conclusion}
\label{sec:conclusion}
\vspace{-0.5\baselineskip}
In this work, we present the first unified model for virtual sensors. Our architecture simultaneously predicts variable sensor subsets, exploiting task synergies while focusing only on relevant input signals to ensure both explainability and efficiency. In our large-scale evaluation on standard and application-specific datasets with billions of samples, we demonstrate substantial efficiency gains, flexible test-time computation trade-offs, and scalability to hundreds of virtual sensors. This enables practical deployment in large sensor networks and on resource-constrained devices. We hope our multi-task model unlocks new virtual sensor applications and contributes to more environmentally sustainable time series modeling.

\textbf{Limitations}\hspace{1.8mm} 
In our work, we do not conduct hyperparameter search for all possible parameters due to the high cost of training large-scale models. We expect even better results with optimized settings.

\section*{Disclaimer}
\vspace{-0.5\baselineskip}
The results, opinions, and conclusions expressed in this publication are not necessarily those of Volkswagen Aktiengesellschaft.

\section*{Funding and competing interests}
\vspace{-0.5\baselineskip}
No funding or competing interests beyond those already disclosed through the authors’ affiliations supported this work.

\small\bibliography{references}

\newpage

\appendix
\crefalias{section}{appendix}
\crefalias{subsection}{appendix}

\section{Positioning of this work}
\label{appendix:positioning}
Our overarching goal is to make virtual sensors scalable to large sensor networks. This positions our work between two previously disjoint literature branches. Classical virtual sensor approaches solve the relevant application problem, but usually train an isolated model with expert-selected inputs for each target sensor. Time series foundation models, in contrast, demonstrate the benefit of shared time series knowledge, but forecast target signals from their own past context. Our work brings the shared-model principle of time series foundation models to the virtual sensor setting, where target signals must be inferred from other measurements rather than extrapolated from their own histories.

\vspace{0.2\baselineskip}
\textbf{Virtual sensor literature}\hspace{1.8mm} Classical virtual sensors typically train an isolated and application-specific model with expert-selected inputs per virtual sensor, limiting scalability. We, in contrast, use one shared backbone for multiple virtual sensors, achieving superior scalability to hundreds of sensors and better prediction quality at comparable inference efficiency. Our approach learns relevant input signal sets, adding explainability without expert selection. Our unified model enables practical virtual sensor deployment in large-scale sensor networks, drastically reducing the training, deployment, and maintenance effort compared to hundreds of isolated models.

\vspace{0.2\baselineskip}
\textbf{Time series foundation model literature}\hspace{1.8mm} 
Time series foundation models demonstrate the benefit of shared time series knowledge across tasks. However, current foundation models rely on in-context learning to forecast future values of signals for which the past context is provided. Virtual sensors solve a different problem: the target signal may be absent from the input context and must be inferred solely from other measurements. Therefore, in-context forecasting alone cannot solve this task, making current time series foundation models incompatible with virtual sensors. We demonstrate that they collapse to zero-prediction in the virtual sensor setting, while our model remains predictive (see below). Our work adapts the unification idea of foundation models to learning multiple virtual sensors in a single model.

\vspace{0.2\baselineskip}
\textbf{Zero-shot generalization}\hspace{1.8mm} Unlike time series foundation models, we do not claim generalization to unseen target sensors, as their past context is absent from the input, hindering zero-shot generalization. Instead, we evaluate the practical case of adding a new virtual sensor via transfer learning. This reaches the performance of training from scratch with only a small fraction of the data and computation in \cref{appendix:transfer_learning}.

\vspace{0.2\baselineskip}
\textbf{Multi-task model}\hspace{1.8mm} We use "multi-task" in the virtual sensing sense that each target virtual sensor is a distinct prediction task with its own output identity and relevant input subset, while all tasks share one backbone. Our unified approach can predict multiple virtual sensors, in contrast to the literature that only focuses on predicting a single output or task. We do not claim that our model handles different task types such as classification, anomaly detection, and imputation. 

\vspace{0.2\baselineskip}
\textbf{Architectural novelty}\hspace{1.8mm} Our backbone architecture deliberately builds on established decoder-only Transformer and patch-tokenization components to isolate our virtual sensor contributions from backbone effects. The novelty of our work is the combination of mechanisms required to make a single model usable for multiple virtual sensors: (i) target-specific prototype tokens and variate embeddings for predicting absent target signals, (ii) sensor selection for flexibly requesting subsets of virtual sensors, (iii) target-specific relevance vectors that learn input signal sets without expert selection, and (iv) structured sparsification that translates learned relevance into runtime and memory savings.

\vspace{0.2\baselineskip}
\textbf{Baseline models}\hspace{1.8mm} LSTMs are common in application-specific virtual sensor literature, making them a relevant practical baseline. We also include transformer baselines with the same backbone as ours to separate backbone effects from the contributions of unification, target-specific prototype tokens and variate embeddings, sensor selection, relevance learning, and structured sparsification. This atomic comparison shows that our performance improvements are not merely due to the transformer architecture.

Here, we additionally compare with time series foundation models in two ways, as shown in \cref{tab:foundation_model_comparison}. Please note that they solve standard forecasting tasks and are not directly applicable to virtual sensors as discussed above. 
(i) In our \textit{virtual sensor} setting, where the target signal's own past is unavailable, all time series foundation models collapse to zero-prediction as expected. This demonstrates that current time series foundation models are incompatible with virtual sensors. In contrast, our model performs well.
(ii) In the \textit{standard forecasting} setting, the foundation models additionally receive the target signal's past context. This is not the virtual sensor setting and gives them an unfair advantage over our model and we report it as a reference. Here, all time series foundation models perform poorly on Solar and generate reasonable forecasts on the other datasets. Even in this setting, our approach achieves superior forecasting performance. \\
Beyond predictive quality, these foundation models lack our efficiency mechanisms for selecting requested target sensors, learning target-specific input sets, and structurally pruning inputs and therefore cannot realize the corresponding runtime and memory savings. \\
Combining these time series foundation models with our proposed mechanisms (prototype tokens, variate embeddings, sensor selection, relevance vectors, and structured sparsification) to transform them from standard forecasting to virtual sensors is theoretically possible but would result in entirely new architectures.

\textbf{Datasets}\hspace{1.8mm} For our evaluation, we use standard time series forecasting datasets with many variates to stress-test our main contributions. Our model has to learn relevant input subsets from a large pool of possible physical sensors, and the output selection mechanism is essential because multiple virtual sensor outputs can be requested at different rates. Traffic, Electricity, and Solar allow us to construct controlled benchmarks with many candidate input signals and multiple virtual sensor outputs, while the CAN bus dataset provides a realistic automotive setting with physically meaningful battery, engine, temperature, torque, current, voltage, and coolant-related virtual sensors. In contrast, common virtual sensor datasets are already small and expert-curated, such as the sideslip-angle dataset \citep{leanza2021sideslip} with \num{13} input signals and a single predefined target. Such datasets are generally useful for single-task virtual sensor models but not for demonstrating the scalability of our multi-task approach.
We nevertheless apply our model to sideslip-angle estimation as a standard virtual sensor task, improving the literature RMSE of \SI{0.57}{\degree} by \SI{5}{\percent} down to \SI{0.54}{\degree}, showing that our approach extends to conventional single-target virtual sensors.

\begin{table}[h]
  \caption{MSE comparison of our unified virtual sensor model with time series foundation models in the virtual sensor setting and the standard forecasting setting.}
  \vspace{0.2\baselineskip}
  \label{tab:foundation_model_comparison}
  \centering
  \resizebox{\textwidth}{!}{
  \begin{tabular}{lccccc@{\hspace{0.8cm}}ccc}
    \toprule
    \multirow{2}{*}{Dataset} & \multicolumn{5}{c}{Virtual sensor} & \multicolumn{3}{c}{Standard forecasting} \\
    \cmidrule(r{0.8cm}){2-6} \cmidrule(l{0.0em}){7-9}
     & Ours & Chronos 2 & TimesFM 2.5 & Toto 2 & Zero-prediction & Chronos 2 & TimesFM 2.5 & Toto 2 \\
    \midrule
    CAN bus     & \textbf{0.121} & \num{0.779} & \num{0.779} & \num{0.779} & \num{0.779} & \num{0.368} & \num{0.304} & \num{0.326} \\
    Traffic     & \textbf{0.282} & \num{1.121} & \num{1.121} & \num{1.121} & \num{1.121} & \num{0.584} & \num{0.634} & \num{0.693} \\
    Electricity & \textbf{0.173} & \num{0.887} & \num{0.887} & \num{0.887} & \num{0.887} & \num{0.221} & \num{0.339} & \num{0.295} \\
    Solar       & \textbf{0.128} & \num{0.967} & \num{0.967} & \num{0.967} & \num{0.967} & \num{1.497} & \num{1.873} & \num{1.595} \\
    \bottomrule
  \end{tabular}
  }
\end{table}

\newpage

\section{Experiments}
\label{appendix:experiments}
In this section, we provide additional information about our experimental settings and resources.

\begin{figure}[h]
        \centering
        \vspace{3pt}
        \begin{subfigure}[b]{5.0in}
        \centering
        \includegraphics[width=5.0in,trim={0in 0.0in 0.0in 0in},clip]{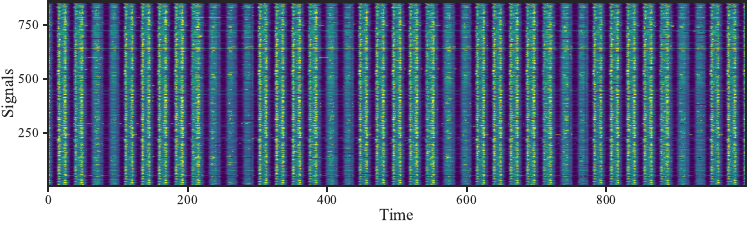}
        \captionsetup{skip=3pt}
        \caption{Traffic}
    \end{subfigure}
    \\
    \vspace{0.4cm}
    \begin{subfigure}[b]{5.0in}
        \centering
        \includegraphics[width=5.0in,trim={0.0in 0.0in 0in 0in},clip]{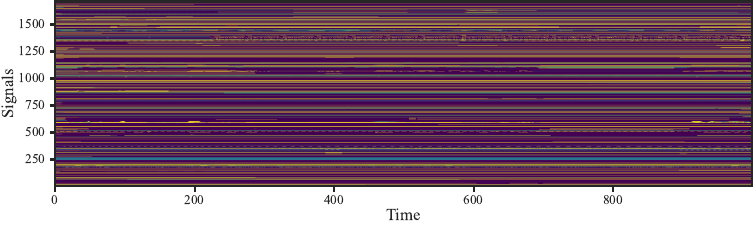}
        \captionsetup{skip=3pt}
        \caption{CAN bus}
    \end{subfigure}
    \\
    \vspace{0.4cm}
    \begin{subfigure}[b]{5.0in}
        \centering
        \includegraphics[width=5.0in,trim={0.0in 0.0in 0in 0in},clip]{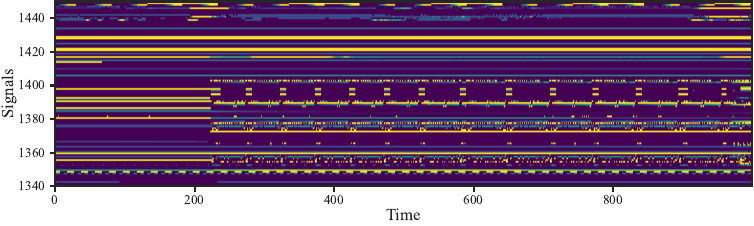}
        \captionsetup{skip=3pt}
        \caption{CAN bus (zoomed)}
    \end{subfigure}
        \vspace{-3pt}
        \caption{Signals on the Traffic and the CAN bus dataset show different temporal characteristics.}
        \label{fig:dataset_structure}
    \end{figure}

\textbf{Datasets}\hspace{1.8mm} Throughout our experiments, we utilize Traffic, Electricity, and Solar as standard benchmark datasets and a CAN bus dataset for evaluating our architecture in more complex and realistic deployment scenarios. Traffic consists of \num{862} sensors measuring the road occupancy in the San Francisco Bay Area every hour and contains roughly \num{15}\,M time series samples. Electricity records the hourly power demand of \num{321} households and consists of approximately \num{8}\,M time series samples. Solar measures power production of \num{137} photovoltaic plants in \num{10} minute intervals with roughly \num{7}\,M time series samples.
Our large-scale CAN bus dataset records \num{1713} time series within a modern electric vehicle. It consists of continuous sensor measurements, categorical states, and event-based signals, in \SI{100}{\milli\second} granularity. Here, each signal has a distinct and interpretable purpose. \\
While time series in the Traffic dataset follow roughly a daily and weekly periodic structure, e.g., due to work traffic, signal interactions in our CAN bus dataset are substantially more complex and diverse. We visualize these different characteristics in \cref{fig:dataset_structure}. Nevertheless, the Traffic dataset still exhibits substantial local heterogeneity across different roads (see \cref{fig:temporal_traffic_1}).

\vspace{0.4cm}

\newpage

\textbf{Virtual sensors}\hspace{1.8mm} We chose our $N=16$ virtual sensors $\mathcal{Z^\prime}$ randomly for the standard time series datasets, including variate IDs $\{74, 88, 113, 173, 280, 311, 312, 352, 355, 478, 539, 646, 650, 715, 775, 776\}$ for Traffic, $\{3, 15, 17, 19, 69, 115, 116, 152, 154, 210, 224, 226, 267, 278, 280, 316\}$ for Electricity, and $\{10, 19, 20, 26, 30, 42, 62, 68, 69, 73, 90, 96, 101, 126, 131, 134\}$ for Solar, as all signals present the same quantity. For the CAN bus dataset with distinct signals, we hand-pick interesting virtual sensor tasks with clear intuitive interpretations according to \cref{tab:virtual_sensor_list}.

  \vspace{-0.5\baselineskip}
\begin{table}[H]
  \caption{Virtual sensors for the CAN bus dataset, which is recorded with an Audi e-tron electric vehicle.}
  \vspace{0.2\baselineskip}
  \label{tab:virtual_sensor_list}
  \centering
  \resizebox{0.35\linewidth}{!}{
  \begin{tabular}{lr}
    \toprule
    Quantity&$j$\\
    \midrule
    \textbf{Battery}&\\
    Battery voltage&\num{1}\\
    Battery current&\num{2}\\
    Battery temperature&\num{3}\\
    Battery coolant outflow temperature&\num{4}\\
    Inverter temperature&\num{5}\\
    \midrule
    \textbf{Engine}\\
    Engine 1 torque&\num{6}\\
    Engine 1 voltage&\num{7}\\
    Engine 1 current&\num{8}\\
    Engine 1 power loss&\num{9}\\
    Engine 1 temperature&\num{10}\\
    Engine 1 rotor temperature&\num{11}\\
    Engine 1 coolant intake temperature&\num{12}\\
    Engine 1 coolant flow rate&\num{13}\\
    \midrule
    \textbf{Vehicle}\\
    Wheel speed (rear, left)&\num{14}\\
    Chassis ground clearance (rear, left)&\num{15}\\
    Air condition power&\num{16}\\
    \bottomrule
  \end{tabular}
  }
\end{table}

\vspace{\baselineskip}

\textbf{Hyperparameters}\hspace{1.8mm} In preliminary experiments, we find promising hyperparameters which we list in \cref{tab:hyperparameters}. We keep these architectural hyperparameters constant throughout our work to ensure a fair and isolated comparison of baselines with models with our proposed mechanisms. 
For all reported results and their underlying models, we conduct an extensive hyperparameter search over optimizer-related parameters, including the learning rate, using HEBO~\citep{CowenRivers2022HEBO}. This way we train competitive models that are robust to varying training dynamics.

  \vspace{-0.5\baselineskip}
\begin{table}[H]
  \caption{Hyperparameters for training our models. We denote sets as $\{\cdots\}$ and continuous search \mbox{spaces as $[\cdots]$.}}
  \vspace{0.2\baselineskip}
  \label{tab:hyperparameters}
  \centering
  \resizebox{0.6\linewidth}{!}{
  \begin{tabular}{lr}
    \toprule
    Hyperparameter & Value  \\
    \midrule
    \textbf{Architectural}&\\
    Patch length&$p=32$\\
    Transformer layers&$L=4$\\
    Attention heads&$4$\\
    Token dimension&$d=512$\\
    Multi-layer perceptron hidden dimension&\num{512}\\
    Activation&ReLU\\
    \midrule
    \textbf{Optimizer-related}&\\
    Seed&\num{2025}\\
    Optimizer&Adam~\citep{kingma2015adam}\\
    Learning rate&$[10^{-6}, 10^{-4}]$\\
    Learning rate warm-up iterations&$\{0, 2, 4\}$\\
    Learning rate warm-up start factor&$[0.3, 0.9]$\\
    Learning rate decay step size&$\{0,1\}$\\
    Learning rate decay gamma&$[0.96, 0.99]$\\
    Dropout&$[0.0, 0.2]$\\
    Batch size&\num{32}\\
    Epochs&\num{80}\\
    Early stopping patience&\num{9}\\
    Loss&MSE\\
    \bottomrule
  \end{tabular}
  }
\end{table}

\textbf{Training stability}\hspace{1.8mm} For our main experiments, we utilize teacher forcing as the standard training strategy and observe stable training dynamics. During our extensive hyperparameter search, we train models with different learning rate and dropout values, which converge to similar MSE, and we find very low sensitivity to these hyperparameters. Specifically, we find MSE standard deviations of \SI{2.5}{\percent} among hyperparameter settings and low correlations of learning rate and MSE of \num{0.00015} and \num{0.077} for dropout. This highlights the robustness and generalization of a single standard training method. In \cref{results:efficient_training_new} we additionally experiment with multi-stage training strategies, further improving MSE.

\vspace{\baselineskip}

\textbf{Reproducibility of measurements}\hspace{1.8mm} 
Regarding \textbf{predictive quality}, we report MSE confidence intervals for our most common experimental settings via test-set bootstrapping here, as repeating all experiments with multiple seeds is computationally infeasible (see computational effort below). \Cref{tab:confidence_intervals} shows tight confidence intervals supporting the significance of our results. \\
For our main experiments, we report efficiency gains in end-to-end \textbf{inference time} and memory, as these real-world measurements are of most practical interest. For runtime profiling, we use the same Nvidia A6000 GPU  with \num{2} warm-up and \num{2} measurement runs per batch to achieve inference time standard deviations $< \SI{2}{\percent}$. Regarding \textbf{memory} allocation, we report dynamic CUDA peak memory, which is a deterministic quantity. Static memory offsets from model parameters are irrelevant in our evaluations as we explore models of similar size throughout our paper. \\
For more in-depth evaluations, we additionally utilize \textbf{sparsity} as a hardware-independent efficiency measure. We explore relations between inference time, memory, and sparsity in \cref{appendix:sparsity}.

\vspace{-0.5\baselineskip}
\begin{table}[H]
  \caption{Bootstrapped test-set MSE confidence intervals.}
  \vspace{0.2\baselineskip}
  \label{tab:confidence_intervals}
  \centering
  \begin{tabular}{lccc}
    \toprule
        Dataset & Mean & \SI{95}{\percent} & \SI{99}{\percent} \\
        \midrule
        CAN bus     & \num{0.136} & [\num{0.122}, \num{0.151}] & [\num{0.118}, \num{0.156}] \\
        Traffic     & \num{0.272} & [\num{0.256}, \num{0.289}] & [\num{0.251}, \num{0.295}] \\
        Electricity & \num{0.174} & [\num{0.169}, \num{0.182}] & [\num{0.167}, \num{0.184}] \\
        Solar       & \num{0.125} & [\num{0.121}, \num{0.129}] & [\num{0.120}, \num{0.130}] \\
        \bottomrule
    \end{tabular}
\end{table}

\vspace{\baselineskip}

\textbf{Computational effort}\hspace{1.8mm} We utilize over \num{48500} compute hours on Nvidia H100 GPUs to train our unified models. To improve training efficiency, we employ \num{16}-bit mixed-precision strategies with \texttt{bfloat16} data types. We estimate the computational effort to reproduce our experiments in \cref{tab:computehours}. Please note that we reuse previously trained models in many of our experiments.

\vspace{-0.5\baselineskip}
\begin{table}[H]
  \caption{Computational effort to reproduce our experiments.}
  \vspace{0.2\baselineskip}
  \label{tab:computehours}
  \centering
  \begin{tabular}{lrr}
    \toprule
    Experiment & GPU hours & Models trained \\
    \midrule
    Main experiment & \num{11384} & \num{1272} \\
    Main experiment -- individual baselines & \num{9392} & \num{1078} \\
    Main experiment -- unified baselines & \num{2044} & \num{492} \\
    Different MSE efficiency trade-offs for variable test-time computation & \num{4242} & \num{160} \\
    Scaling to hundreds of virtual sensors & \num{8776} & \num{660} \\ 
    Explainability of virtual sensors -- Quantitative assessment  & \num{786} & \num{42} \\
    Explainability of virtual sensors -- Synthetic data  & \num{1} & \num{1} \\
    Efficient training strategies -- Training virtual sensors simultaneously & \num{7920} & \num{560} \\
    Efficient training strategies -- Teacher forcing & \num{3964} & \num{36} \\
    \bottomrule
  \end{tabular}
\end{table}

\newpage

\section{Results}
Here, we show additional experiments and results.

\subsection{Main experiments}
\label{appendix:main_experiments}
In \cref{tab_appendix:main_results}, we show additional results of our main experiments on the Electricity and the Solar dataset. On these datasets, our unified approach scales to multiple virtual sensors while substantially reducing computation and memory requirements without affecting predictive quality. 

\vspace{-0.5\baselineskip}
\begin{table}[h]
    \small
      \caption{Comparison of a simple \textbf{unified model} predicting all \num{16} virtual sensors, where we subsequently introduce our proposed mechanisms, with \num{16} \textbf{individual models} predicting a single virtual sensor each. We list MSE, inference time, dynamic CUDA peak memory, and the number of trainable parameters for Electricity and Solar datasets.}
      \vspace{0.2\baselineskip}
      \label{tab_appendix:main_results}
  \centering
  \resizebox{\textwidth}{!}{
  \begin{tabular}{lrrrr@{\hspace{1.0cm}}rrrr}
    \toprule
    \multirow{2}{*}{Architecture} & \multicolumn{4}{@{\hspace{-0.7cm}}c}{Electricity} & \multicolumn{4}{c}{Solar}\\\cmidrule(r{0.95cm}){2-5} \cmidrule(l{0.0em}){6-9}
     & MSE & Time & Mem. & Param. & MSE & Time & Mem. & Param. \\
     \midrule
    \textbf{Individual models}&&&&&&&&\\
    LSTM (literature)  & \textcolor{myred}{\num{0.219}} & \textcolor{mygreen}{\SI{0.31}{\milli\second}} & \textcolor{mygreen}{\SI{0.0009}{\giga\byte}} & \textcolor{myred}{\num{128.3}\,M} & \textcolor{myred}{\num{0.242}} & \textcolor{mygreen}{\SI{0.30}{\milli\second}} & \textcolor{mygreen}{\SI{0.0008}{\giga\byte}} & \textcolor{myred}{\num{121.9}\,M}\\
    Transformer (ours)   & \textcolor{mygreen}{\num{0.145}} & \textcolor{myred}{\SI{2.59}{\milli\second}} & \textcolor{myred}{\SI{0.13}{\giga\byte}} & \textcolor{myred}{\num{115.2}\,M} & \textcolor{mygreen}{\num{0.128}} & \textcolor{myred}{\SI{0.69}{\milli\second}} & \textcolor{myred}{\SI{0.029}{\giga\byte}} & \textcolor{myred}{\num{113.6}\,M}\\
    \midrule
    \textbf{Unified model} & \textcolor{myred}{\num{0.521}} & \textcolor{myred}{\SI{2.63}{\milli\second}} & \textcolor{myred}{\SI{0.13}{\giga\byte}} & \textcolor{mygreen}{\num{7.0}\,M} & \textcolor{myred}{\num{0.149}} & \textcolor{myred}{\SI{0.70}{\milli\second}} & \textcolor{myred}{\SI{0.029}{\giga\byte}} & \textcolor{mygreen}{\num{7.0}\,M}\\
    +\,Variate embedding $\mathcal{V},\mathcal{V^\prime}$ & \textcolor{mygreen}{\num{0.170}} & \textcolor{myred}{\SI{2.64}{\milli\second}} & \textcolor{myred}{\SI{0.13}{\giga\byte}} & \textcolor{mygreen}{\num{7.2}\,M} & \textcolor{mygreen}{\num{0.126}} & \textcolor{myred}{\SI{0.71}{\milli\second}} & \textcolor{myred}{\SI{0.029}{\giga\byte}} & \textcolor{mygreen}{\num{7.1}\,M}\\[0.5em]
    \makecell[l]{+\,Signal relevance $\mathcal{R^\prime}$\\+\,$N_{\mathrm{train}}=4$} & \textcolor{mygreen}{\num{0.174}} & \textcolor{myred}{\SI{2.65}{\milli\second}} & \textcolor{myred}{\SI{0.13}{\giga\byte}} & \textcolor{mygreen}{\num{7.2}\,M} & \textcolor{mygreen}{\num{0.125}} & \textcolor{myred}{\SI{0.72}{\milli\second}} & \textcolor{myred}{\SI{0.029}{\giga\byte}} & \textcolor{mygreen}{\num{7.1}\,M}\\[1em]
    \makecell[l]{+\,Sparse input sets $\mathcal{Z}_j$\\+\,Sensor selection} & \textcolor{mygreen}{\num{0.173}} & \textcolor{mygreen}{\SI{0.31}{\milli\second}} & \textcolor{mygreen}{\SI{0.0089}{\giga\byte}} & \textcolor{mygreen}{\num{7.2}\,M} & \textcolor{mygreen}{\num{0.128}} & \textcolor{mygreen}{\SI{0.14}{\milli\second}} & \textcolor{mygreen}{\SI{0.0036}{\giga\byte}} & \textcolor{mygreen}{\num{7.1}\,M}\\
    \bottomrule
  \end{tabular}
  }
\end{table}

\vspace{2\baselineskip}

\subsection{Different MSE efficiency trade-offs for variable test-time computation}
\label{sec:appendix_mse_efficiency}
\begin{figure}[H]
        \centering
        \vspace{3pt}
        \begin{subfigure}[b]{5.4cm}
        \centering
        \includegraphics[width=5.4cm,trim={0in 0.0in 0.0in 0in},clip]{figures/mask_sparsity_inference/traffic_xlim.pdf}
        \captionsetup{skip=3pt}
        \caption{Traffic}
    \end{subfigure}
    \hspace{0.5cm}
    \begin{subfigure}[b]{5.4cm}
        \centering
        \includegraphics[width=5.4cm,trim={0.0in 0.0in 0in 0in},clip]{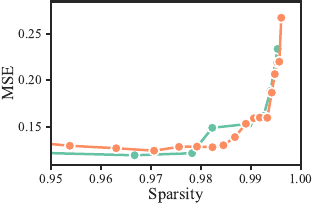}
        \captionsetup{skip=3pt}
        \caption{CAN bus}
    \end{subfigure}
        \vspace{-3pt}
        \caption{Varying our signal importance threshold $r^\prime_{\mathrm{thres}}$ during training or inference generates smooth trade-offs between input set sparsity and MSE on the Traffic and the CAN bus dataset.}
        \label{fig:appendix_mse_efficiency_tradeoff}
    \end{figure}

In \cref{fig:appendix_mse_efficiency_tradeoff} we provide complete results of our investigations of different MSE efficiency trade-offs on the Traffic and the CAN bus dataset.

\vspace{0.2\baselineskip}

\newpage

\subsection{Sensor selection maximizes efficiency without affecting predictive quality}
\label{sec:appendix_sensor_selection}
\begin{figure}[H]
        \centering
        \vspace{3pt}
        \begin{subfigure}[b]{5.4cm}
        \centering
        \includegraphics[width=5.4cm,trim={0in 0.0in 0.0in 0in},clip]{figures/output_selection/traffic.pdf}
        \captionsetup{skip=3pt}
        \caption{Traffic}
    \end{subfigure}
    \hspace{0.5cm}
    \begin{subfigure}[b]{5.4cm}
        \centering
        \includegraphics[width=5.4cm,trim={0.0in 0.0in 0in 0in},clip]{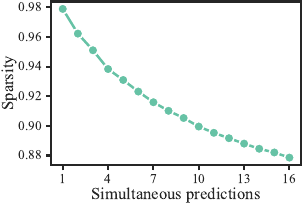}
        \captionsetup{skip=3pt}
        \caption{CAN bus}
    \end{subfigure}
        \vspace{-3pt}
        \caption{Sparsity increases as fewer virtual sensors are predicted simultaneously through our sensor selection mechanism. Results are shown for the Traffic and the CAN bus dataset.}
        \label{fig:appendix_sensor_selection}
    \end{figure}

Here, we demonstrate efficiency gains through our sensor selection mechanism on the Traffic and the CAN bus dataset in \cref{fig:appendix_sensor_selection}.

\vspace{0.2\baselineskip}
\subsection{Efficient training strategies -- Training virtual sensors simultaneously}
\label{sec:appendix_simult_training}
\begin{figure}[H]
        \centering
        \vspace{3pt}
        \begin{subfigure}[b]{5.4cm}
        \centering
        \includegraphics[width=5.4cm,trim={0in 0.0in 0.0in 0in},clip]{figures/per_sample_train/traffic_set_similarity.pdf}
        \captionsetup{skip=3pt}
        \caption{Traffic}
    \end{subfigure}
    \hspace{0.5cm}
    \begin{subfigure}[b]{5.4cm}
        \centering
        \includegraphics[width=5.4cm,trim={0.0in 0.0in 0in 0in},clip]{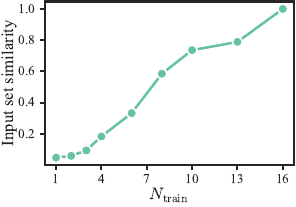}
        \captionsetup{skip=3pt}
        \caption{CAN bus}
    \end{subfigure}
        \vspace{-3pt}
        \caption{Learning different numbers of virtual sensors $N_{\mathrm{train}}$ in a single training iteration influences the similarity of their input signal sets on the Traffic and the CAN bus dataset.}
        \label{fig:appendix_simult_training}
    \end{figure}

\Cref{fig:appendix_simult_training} shows complete results for training multiple virtual sensors simultaneously on the Traffic and the CAN bus dataset.

\newpage

\subsection{Learning relevant input signals}
\label{appendix:learning_input_sets}
\begin{figure}[h]
        \centering
        \vspace{3pt}
        \begin{subfigure}[b]{83.39mm}
        \centering
        \includegraphics[width=83.39mm,trim={0in 0.0in 42.4mm 0in},clip]{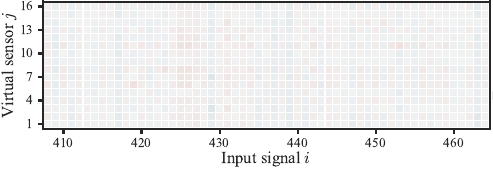}
        \captionsetup{skip=3pt}
        \caption{Signal relevance vectors $r{^\prime_j} \in \mathcal{R^\prime}$, early training}
        \label{fig_appendix:signal_relevance_vect_early_training}
    \end{subfigure}
    \hspace{0.5cm}
    \begin{subfigure}[b]{42.4mm}
        \centering
        \includegraphics[width=42.4mm,trim={83.39mm 0.0in 0in 0in},clip]{figures/signal_relevance_training/mask_and_hist_early.pdf}
        \captionsetup{skip=3pt}
        \caption{Relevance distribution, early}
        \label{fig_appendix:signal_relevance_hist_early_training}
    \end{subfigure}
    \\
    \vspace{0.5cm}
    \begin{subfigure}[b]{83.39mm}
        \centering
        \includegraphics[width=83.39mm,trim={0in 0.0in 42.4mm 0in},clip]{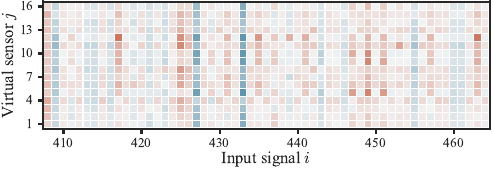}
        \captionsetup{skip=3pt}
        \caption{Signal relevance vectors $r{^\prime_j} \in \mathcal{R^\prime}$, mid training}
        \label{fig_appendix:signal_relevance_vect_mid_training}
    \end{subfigure}
    \hspace{0.5cm}
    \begin{subfigure}[b]{42.4mm}
        \centering
        \includegraphics[width=42.4mm,trim={83.39mm 0.0in 0in 0in},clip]{figures/signal_relevance_training/mask_and_hist_mid.pdf}
        \captionsetup{skip=3pt}
        \caption{Relevance distribution, mid}
        \label{fig_appendix:signal_relevance_hist_mid_training}
    \end{subfigure}
    \\
    \vspace{0.5cm}
            \begin{subfigure}[b]{83.39mm}
        \centering
        \includegraphics[width=83.39mm,trim={0in 0.0in 42.4mm 0in},clip]{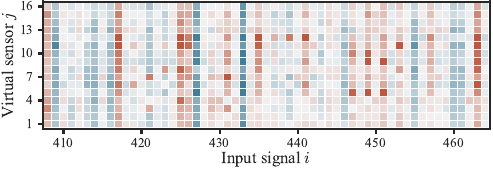}
        \captionsetup{skip=3pt}
        \caption{Signal relevance vectors $r{^\prime_j} \in \mathcal{R^\prime}$, late training}
        \label{fig_appendix:signal_relevance_vect_late_training}
    \end{subfigure}
    \hspace{0.5cm}
    \begin{subfigure}[b]{42.4mm}
        \centering
        \includegraphics[width=42.4mm,trim={83.39mm 0.0in 0in 0in},clip]{figures/signal_relevance_training/mask_and_hist_late.pdf}
        \captionsetup{skip=3pt}
        \caption{Relevance distribution, late}
        \label{fig_appendix:signal_relevance_hist_late_training}
    \end{subfigure}
    \\
    \vspace{0.5cm}
    \begin{subfigure}[b]{83.39mm}
        \centering
        \includegraphics[width=83.39mm,trim={0in 0.0in 42.4mm 0in},clip]{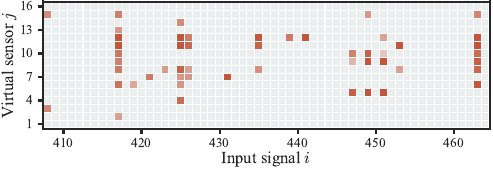}
        \captionsetup{skip=3pt}
        \caption{Signal relevance vectors $r{^\prime_j} \in \mathcal{R^\prime}$, sparse}
        \label{fig_appendix:signal_relevance_vect_sparse}
    \end{subfigure}
    \hspace{0.5cm}
    \begin{subfigure}[b]{42.4mm}
        \centering
        \includegraphics[width=42.4mm,trim={83.39mm 0.0in 0in 0in},clip]{figures/signal_relevance_training/mask_and_hist_dropped.pdf}
        \captionsetup{skip=3pt}
        \caption{Relevance distribution, sparse}
        \label{fig_appendix:signal_relevance_hist_sparse}
    \end{subfigure}
        \caption{\textbf{(a-f)} Signal relevance vectors evolve during training on the CAN bus dataset. \textbf{(g,h)} We sparsify signal relevance using a threshold $r^\prime_{\mathrm{thres}}$ to exclude irrelevant signals, keeping only important ones (arrow).}
    \end{figure}

Our architecture learns individual input signal sets for each virtual sensor as described in \cref{method:input_subsets,sec:input_sparsity}. Here, we provide further insights into the training process. Throughout training, our signal relevance vectors $\mathcal{R^\prime}$ learn signal importance from attention gradients, becoming more diverse as training proceeds in \cref{fig_appendix:signal_relevance_vect_early_training,fig_appendix:signal_relevance_hist_early_training,fig_appendix:signal_relevance_vect_mid_training,fig_appendix:signal_relevance_hist_mid_training,fig_appendix:signal_relevance_vect_late_training,fig_appendix:signal_relevance_hist_late_training}. Next, we distinguish between important and unimportant signals using a threshold $r^\prime_{\mathrm{thres}}$ to sparsify our signal relevance vectors in \cref{fig_appendix:signal_relevance_vect_sparse,fig_appendix:signal_relevance_hist_sparse}, effectively disregarding irrelevant input signals. This finally results in individual input signal sets $\mathcal{Z}_j \subseteq \mathcal{Z}$ for every virtual sensor in \cref{fig_appendix:individual_input_sets}. On the CAN bus dataset, where each signal has a distinct and interpretable purpose, this offers novel explainability of the virtual sensors. Assessment from domain experts further validates meaningful signal selections. 

\vspace{0.4cm}

\begingroup
\makeatletter
\setlength{\@fptop}{0pt}
\setlength{\@fpbot}{0pt plus 1fil}
\makeatother
\newpage
\begin{figure}
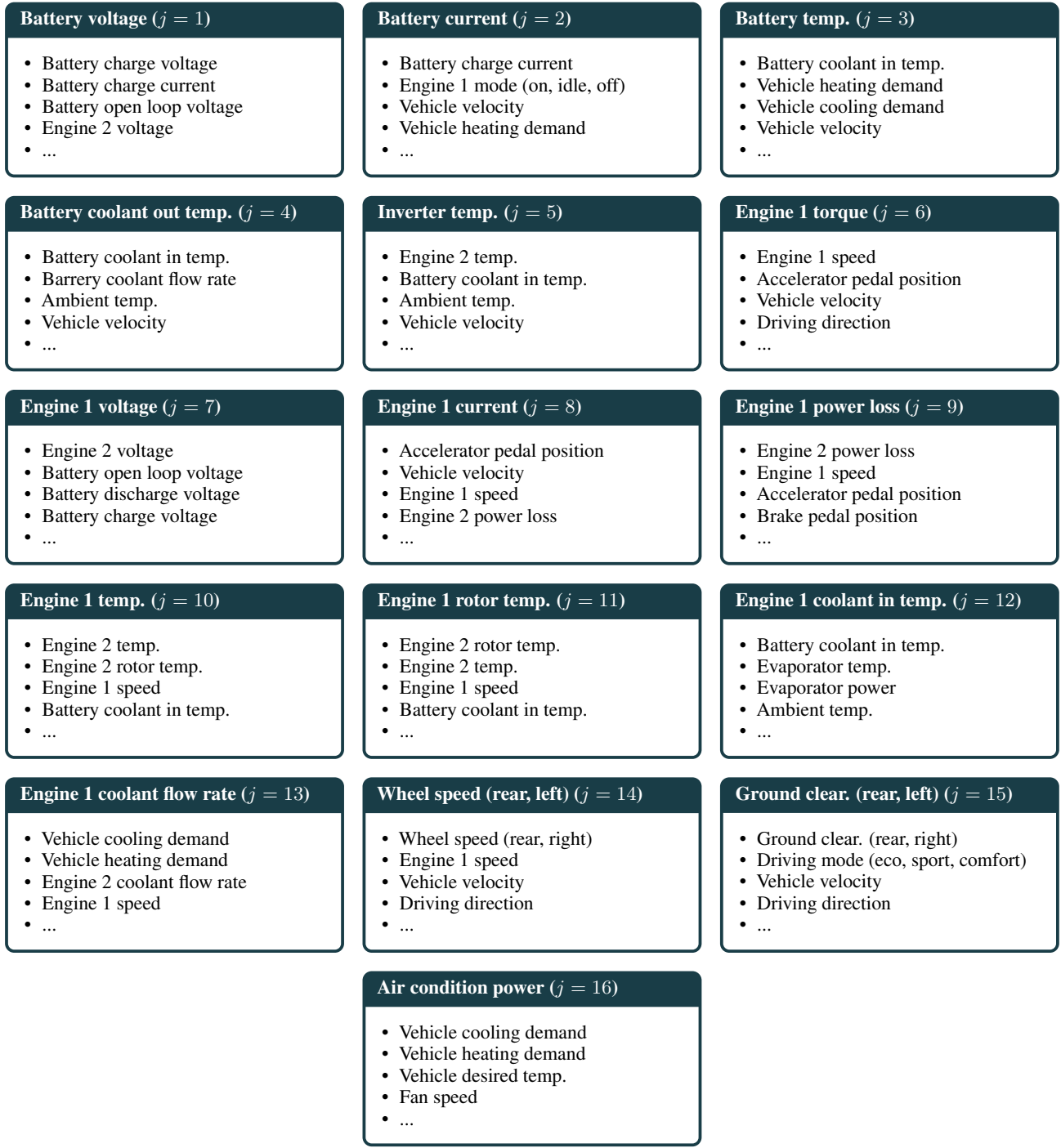

\vspace{0.5\baselineskip}
\centering
\begin{tcbraster}[
  raster columns=3,
  raster equal height,
  raster column skip=2.0mm,
  raster row skip=2.0mm,
  raster left skip=0pt,
  raster right skip=0pt
]
  \begin{tcolorbox}[mycell, title={Battery voltage ($j=1$)}] 
    \begin{itemize}[itemsep=0pt, topsep=0pt, parsep=0pt, partopsep=0pt]
    \item Battery charge voltage
    \item Battery charge current
    \item Battery open loop voltage
    \item Engine 2 voltage
    \item ...
    \end{itemize}
  \end{tcolorbox}
  \begin{tcolorbox}[mycell, title={Battery current ($j=2$)}] 
    \begin{itemize}[itemsep=0pt, topsep=0pt, parsep=0pt, partopsep=0pt]
    \item Battery charge current
    \item Engine 1 mode (on, idle, off)
    \item Vehicle velocity
    \item Vehicle heating demand
    \item ...
    \end{itemize}
  \end{tcolorbox}
  \begin{tcolorbox}[mycell, title={Battery temp. ($j=3$)}] 
    \begin{itemize}[itemsep=0pt, topsep=0pt, parsep=0pt, partopsep=0pt]
    \item Battery coolant in temp.
    \item Vehicle heating demand
    \item Vehicle cooling demand
    \item Vehicle velocity
    \item ...
    \end{itemize}
  \end{tcolorbox}
  \begin{tcolorbox}[mycell, title={\mbox{Battery coolant out temp. ($j=4$)}}] 
    \begin{itemize}[itemsep=0pt, topsep=0pt, parsep=0pt, partopsep=0pt]
    \item Battery coolant in temp.
    \item Battery coolant flow rate
    \item Ambient temp.
    \item Vehicle velocity
    \item ...
    \end{itemize}
  \end{tcolorbox}
  \begin{tcolorbox}[mycell, title={Inverter temp. ($j=5$)}] 
    \begin{itemize}[itemsep=0pt, topsep=0pt, parsep=0pt, partopsep=0pt]
    \item Engine 2 temp.
    \item Battery coolant in temp.
    \item Ambient temp.
    \item Vehicle velocity
    \item ...
    \end{itemize}
  \end{tcolorbox}
  \begin{tcolorbox}[mycell, title={Engine 1 torque ($j=6$)}] 
    \begin{itemize}[itemsep=0pt, topsep=0pt, parsep=0pt, partopsep=0pt]
    \item Engine 1 speed
    \item Accelerator pedal position
    \item Vehicle velocity
    \item Driving direction
    \item ...
    \end{itemize}
  \end{tcolorbox}
  \begin{tcolorbox}[mycell, title={Engine 1 voltage ($j=7$)}] 
    \begin{itemize}[itemsep=0pt, topsep=0pt, parsep=0pt, partopsep=0pt]
    \item Engine 2 voltage
    \item Battery open loop voltage
    \item Battery discharge voltage
    \item Battery charge voltage
    \item ...
    \end{itemize}
  \end{tcolorbox}
  \begin{tcolorbox}[mycell, title={Engine 1 current ($j=8$)}] 
    \begin{itemize}[itemsep=0pt, topsep=0pt, parsep=0pt, partopsep=0pt]
    \item Accelerator pedal position
    \item Vehicle velocity
    \item Engine 1 speed
    \item Engine 2 power loss
    \item ...
    \end{itemize}
  \end{tcolorbox}
  \begin{tcolorbox}[mycell, title={Engine 1 power loss ($j=9$)}] 
    \begin{itemize}[itemsep=0pt, topsep=0pt, parsep=0pt, partopsep=0pt]
    \item Engine 2 power loss
    \item Engine 1 speed
    \item Accelerator pedal position
    \item Brake pedal position
    \item ...
    \end{itemize}
  \end{tcolorbox}
  \begin{tcolorbox}[mycell, title={Engine 1 temp. ($j=10$)}] 
    \begin{itemize}[itemsep=0pt, topsep=0pt, parsep=0pt, partopsep=0pt]
    \item Engine 2 temp.
    \item Engine 2 rotor temp.
    \item Engine 1 speed
    \item Battery coolant in temp.
    \item ...
    \end{itemize}
  \end{tcolorbox}
  \begin{tcolorbox}[mycell, title={Engine 1 rotor temp. ($j=11$)}] 
    \begin{itemize}[itemsep=0pt, topsep=0pt, parsep=0pt, partopsep=0pt]
    \item Engine 2 rotor temp.
    \item Engine 2 temp.
    \item Engine 1 speed
    \item Battery coolant in temp.
    \item ...
    \end{itemize}
  \end{tcolorbox}
  \begin{tcolorbox}[mycell, title={\mbox{Engine 1 coolant in temp. ($j=12$)}}] 
    \begin{itemize}[itemsep=0pt, topsep=0pt, parsep=0pt, partopsep=0pt]
    \item Battery coolant in temp.
    \item Evaporator temp.
    \item Evaporator power
    \item Ambient temp.
    \item ...
    \end{itemize}
  \end{tcolorbox}
  \begin{tcolorbox}[mycell, title={\mbox{Engine 1 coolant flow rate ($j=13$)}}] 
    \begin{itemize}[itemsep=0pt, topsep=0pt, parsep=0pt, partopsep=0pt]
    \item Vehicle cooling demand
    \item Vehicle heating demand
    \item Engine 2 coolant flow rate
    \item Engine 1 speed
    \item ...
    \end{itemize}
  \end{tcolorbox}
  \begin{tcolorbox}[mycell, title={\mbox{Wheel speed (rear, left) ($j=14$)}}] 
    \begin{itemize}[itemsep=0pt, topsep=0pt, parsep=0pt, partopsep=0pt]
    \item Wheel speed (rear, right)
    \item Engine 1 speed
    \item Vehicle velocity
    \item Driving direction
    \item ...
    \end{itemize}
  \end{tcolorbox}
  \begin{tcolorbox}[mycell, title={\mbox{Ground clear. (rear, left) ($j=15$)}}] 
    \begin{itemize}[itemsep=0pt, topsep=0pt, parsep=0pt, partopsep=0pt]
    \item Ground clear. (rear, right)
    \item Driving mode (eco, sport, ...)
    \item Vehicle velocity
    \item Driving direction
    \item ...
    \end{itemize}
  \end{tcolorbox}
  \center
  \begin{tcolorbox}[mycell, title={Air condition power ($j=16$)}] 
    \begin{itemize}[itemsep=0pt, topsep=0pt, parsep=0pt, partopsep=0pt]
    \item Vehicle cooling demand
    \item Vehicle heating demand
    \item Vehicle desired temp.
    \item Fan speed
    \item ...
    \end{itemize}
  \end{tcolorbox}
\end{tcbraster}
\vspace{0.2\baselineskip}
\caption{Individual input signal sets $\mathcal{Z}_j \subseteq \mathcal{Z}$ for each virtual sensor $j$ for the CAN bus dataset, learned by our model. For clarity, we list only \num{4} input signals, while virtual sensors use \num{38} on average. Further, heating and cooling demand signals relate to the vehicle's interior but also to drivetrain components such as the battery and engine. (See \cref{tab:virtual_sensor_list} for non-abbreviated virtual sensor names.)}
\label{fig_appendix:individual_input_sets}
\end{figure}
\clearpage
\endgroup

\subsection{Interactions}
\label{sec_appendix:interactions}
\begin{figure}[h]
        \centering
        \vspace{3pt}
        \begin{subfigure}[b]{5.4cm}
        \centering
        \includegraphics[width=5.4cm,trim={0in 0.0in 0.0in 0in},clip]{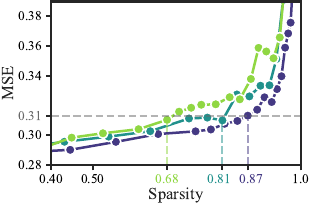}
        \captionsetup{skip=3pt}
        \caption{Prediction quality and sparsity}
        \label{fig_appendix:interactions_a}
    \end{subfigure}
    \hspace{0.5cm}
    \begin{subfigure}[b]{5.4cm}
        \centering
        \includegraphics[width=5.4cm,trim={0.0in 0.0in 0in 0in},clip]{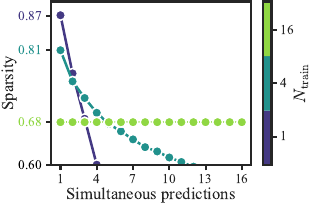}
        \captionsetup{skip=3pt}
        \caption{Prediction of multiple virtual sensors}
        \label{fig_appendix:interactions_b}
    \end{subfigure}
        \vspace{-3pt}
        \caption{Learning different numbers of virtual sensors $N_{\mathrm{train}}$ in every training iteration influences \textbf{(a)} tradeoffs between predictive quality and sparsity and \textbf{(b)} inference efficiency when predicting multiple virtual sensors simultaneously on the Traffic dataset.}
    \end{figure}

There are complex interactions between training efficiency, inference efficiency, and prediction quality, which we explore here in detail. Increasing $N_{\mathrm{train}}$ to learn multiple virtual sensors at a time improves training efficiency. It also leads to more similar input sets $\mathcal{Z}_j \subseteq \mathcal{Z}$ for the virtual sensors (see \cref{fig:tv_per_sample_train_traffic}). This has two major effects during inference. First, for a fixed prediction quality, the sensor's input sets need to be larger and less sparse as they are less specialized (see \cref{fig_appendix:interactions_a}). This reduces peak efficiency when predicting virtual sensors individually \cref{fig_appendix:interactions_b}. Second, when forecasting multiple virtual sensors at a time, more similar input sets lead to graceful scaling as in \cref{fig_appendix:interactions_b}. Here, synergies among input sets benefit efficiency. \\
For the Traffic dataset, peak inference efficiency for individual virtual sensors is \SI{3.8}{\times} higher for runtime and \SI{5.2}{\times} for memory compared to constant scaling behavior for multiple virtual sensors. 
Generally, we suggest training multiple virtual sensors simultaneously for multi-sensor inference use cases, while learning single virtual sensors at a time for peak inference efficiency of individual virtual sensors. Note that quantitative efficiency gains are strongly dependent on the choice of virtual sensors and the dataset.

\vspace{\baselineskip}
\subsection{Real-world efficiency gains through sparsity}
\label{appendix:sparsity}
\begin{figure}[h]
        \centering
        \vspace{3pt}
        \begin{subfigure}[b]{5.4cm}
        \centering
        \includegraphics[width=5.4cm,trim={0in 0.0in 0.0in 0in},clip]{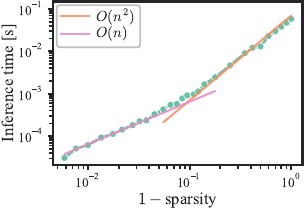}
        \captionsetup{skip=3pt}
        \caption{Inference time}
    \end{subfigure}
    \hspace{0.5cm}
    \begin{subfigure}[b]{5.4cm}
        \centering
        \includegraphics[width=5.4cm,trim={0.0in 0.0in 0in 0in},clip]{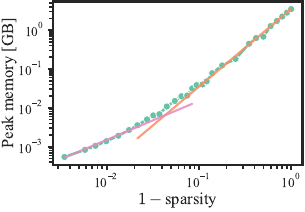}
        \captionsetup{skip=3pt}
        \caption{Memory}
    \end{subfigure}
        \vspace{-3pt}
        \caption{Structured sparsity induces savings in inference time and dynamic CUDA peak memory. Note that we visualize $1-\text{sparsity}$ on logarithmic axes. Therefore, quadratic and linear complexities transform into linear functions with slopes of \num{2} and \num{1}, respectively.}
        \label{fig:sparsity_efficiency_gains}
    \end{figure}

In our main experiments, we report efficiency gains in inference time and memory, which are of high practical relevance. For more in-depth analysis, we also report sparsity. Here, we explore relations between sparsity, runtime, and memory allocation in detail. \\
Our architecture employs sparsity in a structured way by reducing the transformer's input tokens $s$ to a minimum. It focuses only on relevant input signals for a requested set of virtual sensors as described in \cref{sec:method_output_selection,method:input_subsets,sec:input_sparsity,sec:output_selection}. We profile our architecture with varying sparsity values in the following. Remarkably, our structured sparsity directly yields real-world savings in computation time and memory over several orders of magnitude, as our results in \cref{fig:sparsity_efficiency_gains} show. For low sparsity values, the attention operation with quadratic complexity is most dominant, while the transformer's multi-layer perceptron with linear complexity is the limiting operation in high-sparsity regimes.

\newpage

\subsection{Visualization of virtual sensor predictions}
\label{appendix:visualization}

\begin{figure}[h]
        \centering
        \vspace{3pt}
        \begin{subfigure}[b]{2.5in}
        \centering
        \includegraphics[width=2.5in,trim={0in 0.0in 0.0in 0in},clip]{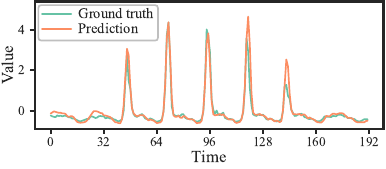}
        \captionsetup{skip=3pt}
        \caption{Virtual sensor $j=1$ (Traffic variate \num{74})}
    \end{subfigure}
    \hspace{0.8cm}
    \begin{subfigure}[b]{2.5in}
        \centering
        \includegraphics[width=2.5in,trim={0.0in 0.0in 0in 0in},clip]{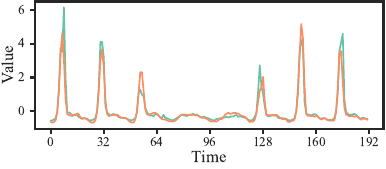}
        \captionsetup{skip=3pt}
        \caption{Virtual sensor $j=1$ (Traffic variate \num{74})}
    \end{subfigure}
    \\
    \vspace{0.4cm}
    \begin{subfigure}[b]{2.5in}
        \centering
        \includegraphics[width=2.5in,trim={0.0in 0.0in 0in 0in},clip]{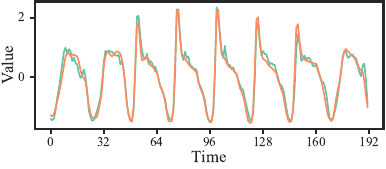}
        \captionsetup{skip=3pt}
        \caption{Virtual sensor $j=3$ (Traffic variate \num{113})}
    \end{subfigure}
    \hspace{0.8cm}
    \begin{subfigure}[b]{2.5in}
        \centering
        \includegraphics[width=2.5in,trim={0.0in 0.0in 0in 0in},clip]{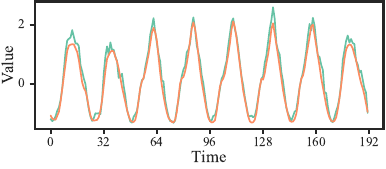}
        \captionsetup{skip=3pt}
        \caption{Virtual sensor $j=4$ (Traffic variate \num{173})}
    \end{subfigure}
    \\
    \vspace{0.4cm}
    \begin{subfigure}[b]{2.5in}
        \centering
        \includegraphics[width=2.5in,trim={0.0in 0.0in 0in 0in},clip]{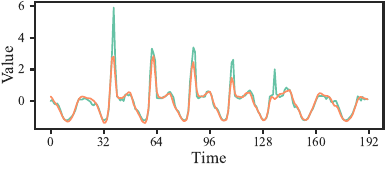}
        \captionsetup{skip=3pt}
        \caption{Virtual sensor $j=6$ (Traffic variate \num{311})}
    \end{subfigure}
    \hspace{0.8cm}
    \begin{subfigure}[b]{2.5in}
        \centering
        \includegraphics[width=2.5in,trim={0.0in 0.0in 0in 0in},clip]{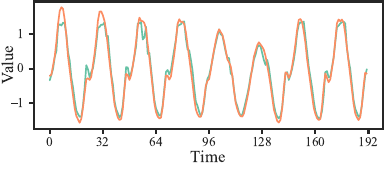}
        \captionsetup{skip=3pt}
        \caption{Virtual sensor $j=12$ (Traffic variate \num{646})}
    \end{subfigure}
    \\
    \vspace{0.4cm}
    \begin{subfigure}[b]{2.5in}
        \centering
        \includegraphics[width=2.5in,trim={0.0in 0.0in 0in 0in},clip]{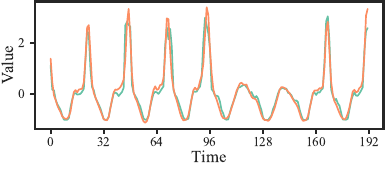}
        \captionsetup{skip=3pt}
        \caption{Virtual sensor $j=14$ (Traffic variate \num{715})}
    \end{subfigure}
    \hspace{0.8cm}
    \begin{subfigure}[b]{2.5in}
        \centering
        \includegraphics[width=2.5in,trim={0.0in 0.0in 0in 0in},clip]{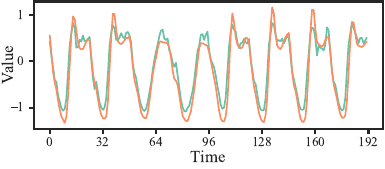}
        \captionsetup{skip=3pt}
        \caption{Virtual sensor $j=15$ (Traffic variate \num{775})}
    \end{subfigure}
        \caption{Virtual sensor predictions on the Traffic dataset. Our sensor selection mechanism successfully switches between virtual sensors with different temporal patterns.}
        \label{fig:temporal_traffic_1}
    \end{figure}

\begin{figure}[h]
\vspace{0.25\baselineskip}
    \centering
    \includegraphics[width=5.4cm,trim={0in 0.0in 0in 0in},clip]{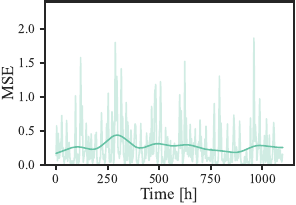}
    \vspace{-3pt}
    \caption{MSE remains stable for very long virtual sensor operation on the Traffic dataset.}
    \label{fig_appendix:long_run_traffic}
    \vspace{\baselineskip}
\end{figure}

In this section, we illustrate predictions of our unified virtual sensor model on the Traffic dataset. In contrast to standard time series forecasting literature \citep{nie2023patchtst}, our architecture does not have a fixed context horizon. Instead, it forecasts virtual sensors $p=32$ time steps from other measurement signals, starting from empty prototype tokens. In the following autoregressive cycles, our model also utilizes its past virtual sensor predictions. Over time, past context grows, mimicking practical deployment scenarios of virtual sensors. \\
Our visualizations in \cref{fig:temporal_traffic_1} demonstrate that our architecture successfully predicts initial virtual sensor values from empty prototypes. Subsequent autoregressive cycles do not show error accumulation and predicted patches connect smoothly. Even over long-horizon rollouts exceeding \num{1000} hours of traffic flow, the predictive loss remains stable in \cref{fig_appendix:long_run_traffic}. Finally, our sensor selection mechanism successfully switches between multiple virtual sensors with different temporal characteristics, measuring traffic flow at different locations in San Francisco. All this visual evidence aligns with our previous results, demonstrating the flexibility of our approach.

\vspace{0.4cm}

\newpage

\subsection{Transfer learning}
\label{appendix:transfer_learning}
\begin{figure}[h]
\vspace{0.25\baselineskip}
    \centering
    \includegraphics[width=5.4cm,trim={0in 0.0in 0in 0in},clip]{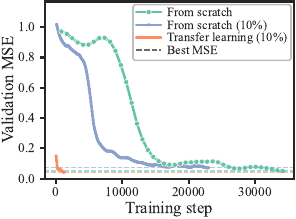}
    \vspace{-3pt}
    \caption{Comparison of training from scratch with full and limited (\SI{10}{\percent}) data to transfer learning for introducing a new virtual sensor on the CAN bus dataset under constrained data availability. Note that validation MSE is reported at the end of each epoch -- this is why initial transfer learning loss is already low.}
    \label{fig_appendix:transfer_learning}
\end{figure}

In practical applications, data availability is often limited, and new virtual sensors may be required over the lifetime of a model. Time series foundation models offer appealing properties, including zero‑shot generalization to unseen datasets and transfer learning. While comparable zero‑shot generalization is infeasible for virtual sensors\footnote{Zero‑shot generalization in time series foundation models relies on in‑context learning, i.e., identifying patterns in the input series and extrapolating them as predictions. Virtual sensors, however, correspond to new target signals that are not present in the input context, precluding in‑context learning.} and our scope is to learn multiple sensors within a single network, we demonstrate that transfer learning allows new virtual sensors to be efficiently introduced into an already trained model using only limited data. \\
To investigate the transfer learning capabilities of our architecture, we first train a model on the CAN bus dataset with $N=15$ virtual sensors, excluding the wheel speed sensor ($j=14$). In a second step, we introduce the wheel speed virtual sensor via transfer learning using only \SI{10}{\percent} of the original training data. \Cref{fig_appendix:transfer_learning} shows that this approach enables the efficient integration of a new virtual sensor into an existing model under constrained data availability. Transfer learning achieves the same predictive performance as training from scratch with full data while requiring only \SI{3.6}{\percent} of the training iterations. Moreover, compared to training from scratch under limited data, transfer learning yields models with improved predictive accuracy. These results indicate that transfer learning exploits synergies and previously acquired system knowledge to efficiently integrate new virtual sensors.

\end{document}